\documentclass[a4paper,12pt,oneside]{amsart}
\usepackage{authblk}

\usepackage{geometry}
\usepackage{epstopdf}
\usepackage{mathtools}
\usepackage[cmtip,all]{xy}

\usepackage{graphicx}
\usepackage{amssymb}
\usepackage{amsfonts}
\usepackage{amsmath}
\usepackage{mathtools}
\usepackage{epstopdf}
\usepackage{color}
\usepackage[all,cmtip]{xy}
\usepackage{array}
\usepackage{tikz-cd}
\usepackage{tabularx}
\usepackage{fancyhdr}
\usepackage{booktabs}
\usepackage[normalem]{ulem}

\usepackage{caption}
\usepackage{subcaption}

\usepackage{lipsum}

\usepackage{bm,dsfont}
\usepackage{multirow}
\usepackage{enumitem}
\usepackage{color}
\definecolor{red}{rgb}{0.70,0.13,0.13}
\definecolor{green}{rgb}{0.13,0.55,0.13}
\definecolor{blue}{rgb}{0.25, 0.41, 0.88}
\usepackage{hyperref}
\hypersetup{
colorlinks = true,
linkcolor = blue,
citecolor = green,
urlcolor = blue}

\newcommand{\dd}{\mathrm{d}}

\newcommand{\dsE}{\mathbb{E}}

\newcommand{\dsN}{\mathbb{N}}

\newcommand{\dsR}{\mathbb{R}}

\newcommand{\scD}{\mathcal{D}}

\newcommand{\scG}{\mathcal{G}}

\newcommand{\scL}{\mathcal{L}}

\newcommand{\scR}{\mathcal{R}}

\newcommand{\Zeta}{\mathrm{Z}}

\newcommand{\Tr}{\operatorname{Tr}}

\newcommand{\eqnref}[1]{Eq.\,\eqref{#1}}
\newcommand{\figref}[1]{Fig.\,\ref{#1}}
\newcommand{\tabref}[1]{Tab.\,\ref{#1}}

\usepackage{times} 

\usepackage[english]{babel}
\usepackage[numbers,sort&compress]{natbib}

\theoremstyle{plain}

\newtheorem{thm}{Theorem}[section]

\newtheorem{rmk}[thm]{Remark}

\newtheorem*{prop*}{Proposition}

\newtheorem{prop-defi}[thm]{Proposition-Definition}
\newtheorem{thm-defi}[thm]{Theorem-Definition}
\newtheorem{lem-defi}[thm]{lema-Definition}

\newtheorem{exam}[thm]{Example}

\newenvironment{itemize*}%
  {\vspace{-5pt}\begin{itemize}%
    \setlength{\itemsep}{0pt}%
    \setlength{\parskip}{0pt}}%
  {\end{itemize}\vspace{-5pt}}
 \newenvironment{enumerate*}%
  {\vspace{-5pt}\begin{enumerate}[label={(\arabic*)}]%
    \setlength{\itemsep}{0pt}%
    \setlength{\parskip}{0pt}}%
  {\end{enumerate}\vspace{-5pt}}
\usepackage{tikz-cd}

\begin{document}

\title[CAL and RG operators for classifiers]{Categorical Representation Learning and RG flow operators for algorithmic classifiers}
\maketitle

\begin{center}
\author{}{Artan Sheshmani${}^{1,2,3, 5,6}$ and Yi-Zhuang You${}^{4,5}$ and Wenbo Fu${}^{5}$ and Ahmadreza Azizi${}^{5}$}
\end{center}
\address{${}^1$  Center for Mathematical Sciences and\\ Applications, Harvard University, Department of Mathematics, and Harvard University Physics department, Jefferson Laboratory, 17 Oxford St, Cambridge, MA 02138}
\address{${}^2$ IMSA, University of Miami}
\address{${}^3$ National Research University Higher School of Economics, Russian Federation, Laboratory of Mirror Symmetry, NRU HSE, 6 Usacheva str.,Moscow, Russia, 119048}
\address{${}^4$ University of California San Diego, Department of Physics, Condensed matter group, UC San Diego 9500 Gilman Dr. La Jolla, CA 92093}
\address{${}^5$ QGNai INC. (Quantum Geometric networks for artificial intelligence), 83 Cambridge Parkway, Unit W806. Cambridge, MA, 02142}
\address{${}^6$ NSF AI Institute for Artificial Intelligence and Fundamental Interactions}
\date{\today}

\begin{abstract}Following the earlier formalism of the categorical representation learning \cite{categorifier} by the first two authors, we discuss the construction of the ``RG-flow based categorifier". Borrowing ideas from theory of renormalization group flows (RG) in quantum field theory, holographic duality, and hyperbolic geometry, and mixing them with neural ODE's, we construct a new algorithmic natural language processing (NLP) architecture, called the RG-flow categorifier or for short the RG categorifier, which is capable of data classification and generation in all layers. We apply our algorithmic platform to biomedical data sets and show its performance in the field of sequence-to-function mapping. In particular we apply the RG categorifier to particular genomic sequences of flu viruses and show how our technology is capable of extracting the information from given genomic sequences, find their hidden symmetries and dominant features, classify them and use the trained data to make stochastic prediction of new plausible generated sequences associated with new set of viruses which could avoid the human immune system. The content of the current article is part of the recent US patent application submitted by first two authors (U.S. Patent Application No.: 63/313.504). 
\smallskip



\end{abstract}
\tableofcontents

\section{Introduction}

The renormalization group (RG) \cite{Polchinski1984Renormalization} is a powerful and useful set of methods developed in statistical physics and quantum field theory to deal with many-body problems. It helps the physicists to establish the connection between the microscopic laws of physics and the macroscopic collective behaviors of the system. It starts with a many-body system at the microscopic scale, and then performs the coarse graining iteratively to group the fundamental building blocks together into larger and larger clusters. Meanwhile, it constructs the effective descriptions of the clusters at each different scale and extracts the effective interaction among them. At the end, the many-body system can be reduced to a few-body system at the highest scales, which enables the understanding of complex systems and their collective behaviors at large scale.

This idea can be particularly useful for representation learning and classification tasks in machine learning. There are many examples of many-body systems in machine learning tasks. For instance, an image can be viewed as a system of many pixels, and a sequence can be viewed as a system of many tokens. It is desired to see whether the idea of renormalization group can also be applied to extract the overall representations of images and sequences from their microscopic representations.

In terms of mathematics, the existence of profound connections between quantum field theory and geometry/ topology has been a source of many exciting research activities. One of them, as an example, is the interesting connection between theory of RG flows, for a particular set of quantum field theories in physics, and the geometric theory of Ricci flows in mathematics. The theory of Ricci flows was developed by Richard Hamilton in the 80's \cite{Hamilton1, Hamilton2, Hamilton3, Hamilton4, Hamilton5}.  Given  a smooth manifold, $M$, a Riemannian metric, $g$, on $M$ defines a bilinear positive-definite product on tangent space, $T_{p}M$, for each point $p\in M$. This bilinear form is a 2-tensor which locally in an open neighborhood $U\subset M$ of $p$, will have a matrix representation. One can then investigate whether infinitesimal deformations of the metric on $M$ would provide interesting information about its geometry or topology. For instance given a 1-parameter family, $g_{t}, t\in (a,b)$ of metrics on $M$, one can study the variation of $g$ with respect to the parameter $t$. The derivative $\displaystyle{\frac{\partial g_{t}}{\partial t}}$ will then provide for every fixed choice of $t$ and fixed point $p$ a bilinear inner product form (i.e. a 2-tensor) on $T_{p}M$. It turns out that variation of metric in a 1-parameter family provides one with a differential equation $$\displaystyle{\frac{\partial g_{t}}{\partial t}}=-2\text{Ric}^{g_{t}},$$ where the term on the right hand side is the Ricci curvature tensor, $\text{Ric}^{g_{t}}$, named after  Gregorio Ricci-Curbastro, measuring that, how for each fixed choice of $g_{t}$, the geometry of space is curved as one moves along the geodesics on the manifold $M$. \\

The connection between RG flows for nonlinear sigma models in physics and the Ricci flow for Riemannian manifolds in mathematics is quite known for a while, since the earlier work of Daniel Freidan \cite{Freidan}, Zamalodchikov \cite{Zomolodchikov}, Tseytlin \cite{Tseytlin}, as well as ground breaking work of Gregory Perelman \cite{Perelman} in proof of Poincare conjecture, and more recently Carfora \cite{Carfora}. In he next section we briefly provide an expository account of RG flows in the context of Ricci geometry following the work of Carfora \cite{Carfora}. It must be noted that our focus in the current article is to implement RG flows for developing algorithmic architectures in mathematical artificial intelligence, therefore later we quickly diverge from its connection to Ricci geometry, and focus solely on RG networks. We encourage the interested readers to study the resources provided above to gain a deeper understanding of the connections between the two frameworks in physics and mathematics.\\
 
When it comes to implementing RG flow theory in machine learning, the key challenge lies in the difficulty in constructing the coarse graining transformation at each RG step. In physics, the RG rules are usually specified by human, such as the majority vote in real-space RG or the momentum-shell integration in field theoretic RG. These intuitions may not be immediately applicable to realistic dataset of images and sequences, as the underlying coarse graining rules may be much more complicated compared to physics systems. This calls for machine learning methods to enable algorithm to design and optimize the RG transformation in adaptation to the given dataset. One important idea is borrowed from the holographic duality in physics, which states that the RG transformation can  be viewed as a holographic mapping of a field configuration from a flat (boundary) space to a hyperbolic (bulk) space with one-higher dimension, such that the long-range correlation in the original field configuration can be equivalently represented as short-range correlation in the bulk space. So the optimal RG transformation can be defined as a bijective holographic mapping that disentangles the features at different hierarchies as much as possible. This allows us to embed the bijective holographic map in the flow-based generative model, and use the unsupervised machine learning technique to train the optimal RG transformation. This idea is first proposed in Ref.\,\cite{Li2018Neural} and further developed in later works \cite{Hu2020Machine,Hu2020RG-Flow}. The current article further develops the machine-learning RG method by combining the RG-flow model with neural ODE techniques \cite{Chen2018Neural}, and explores its application to representation learning of sequential data.

\section*{Acknowledgements}
The first author would like to aknowledge support by National Science Foundation SBRI grant No: 2109928, as well as support by the National Science Foundation under Cooperative Agreement PHY-2019786 (the NSF AI Institute for Artificial Intelligence and Fundamental Interactions, http://iaifi.org/).  The second author was supported by a startup fund provided by UCSD and the UC Hellman fellowship. We acknowledge the discussion with Hong-Ye Hu. 
\section{Analytic construction of RG flow operator on moduli space of smooth maps}
\subsection{A special example of our construction, using Laplacians and curvature form}
A large part of current section is based on work of Carfora \cite{Carfora} in relating RG flows for a specific set of QFT's and the Ricci flow construction for Riemannian manifolds. Moreover, the content in this section owes its existence to another highly recommended source, specially for a working mathematician, that is the outstanding work of Kevin Costello \cite{Costello} in his mathematical formulation of perturbative quantum field theory.

For the time being, we use the introduction to a geometric construction of the RG flow, outlined below, as it is suitably intuitive, pleasantly elegant, and mainly since it will provide us later with the shortest pathways to generalize our constructions in several ways, for instance: by deviating from the classical setup, via altering (and generalizing) our action integrals made with Laplacians and curvature forms to more general actions, or by altering the base geometrical spaces from smooth manifolds to non-smooth algebraic varieties, or discrete lattices. \\

Let $C, X$ denote respectively a compact oriented Riemann surface and a compact oriented smooth manifold of dimension at least 2, both equipped with a Riemannian metric, and defined over a base number field $\mathbb{K}$.  Let $\text{Map}(C, X):=\{f: C\to X\}$ be the associated space of all continuous maps from domain $C$ to $X$. The construction of RG flow is based on considering a family of Lagrangians $\mathcal{L}(f, \phi_{i}, i=1,\cdots, n)$ associated to this space, defined as a morphism $$\mathcal{L}(f, \phi_{i}, i=1,\cdots, n):= \text{Map}(C, X)\times H^{*}(X, \mathbb{K})^{\otimes n} \to C^{\infty}(C,X)$$ taking a tuple of fields $(f, \phi_{1},\cdots, \phi_{n})$ on $X$ to the space of smooth integrable functions on $X$. Note that here the notatation $H^{*}(X, \mathbb{K})^{\otimes n}$ means that  the fields $\phi_{i}, i=1, \cdots, n$ are realized as sections of a sheaf of differentially graded algebras over $X$, sitting in appropriate cohomological degrees on $X$. Moreover, we require the Lagrangians to be invariant under the action of diffeomorphism groups, $\mathcal{D}\textit{iff}(C), \mathcal{D}\textit{iff}(X)$ on $C$ and $X$ respectively.\\
Integrating the Lagrangian over the associated domain Riemann surface induces the Lagrangian action integral $$\int_{C}\mathcal{L}(f, \phi_{i}, i=1,\cdots, n).$$Let the metric tensors on $C$ and $X$ be respectively denoted by $\mu_{mn}, m,n=1,2$ and $g_{ij}, i,j=1,\cdots n$. Suppose that the local coordinates on $C$ are given by $x$ (that is $x:=(x_{1}, x_{2})$). Then a typical form of  such Lagrangian action integral as defined above is given as
\begin{align}\label{eq:Laplacian}
&\int_{C}\mathcal{L}(f, \phi_{i}, i=1,\cdots, n)=\int_{C} \lambda^{-1} \left[\mu_{mn}(x)\partial_{m} f_{i}(x)\partial_{n} f_{j}(x)g_{ij}(f(x))+ \lambda \rho(f)\mathcal{K}\right]d\nu_{C}\notag\\
& m,n=1,2\,\,\,\, \text{and}\,\,\,\, i,j=1,\cdots, n.\notag\\
\end{align}
where $\lambda$ is a coupling parameter, $\nu_{C}$ is a measure on $C$, $\rho: X\to \mathbb{K}\in C^{\infty}(X)$ is a smooth function on $X$, and $\mathcal{K}$ is the Gaussian curvature on $C$ with respect to the metric $\mu$. Here the fields associated to the Lagrangian action integral are given as $$\phi=\lambda^{-1}(g, \lambda \rho).$$
\begin{rmk}
By this notation we mean that the coupling constant $\lambda$ has dependence on parameters $g, \rho$.
\end{rmk}
\begin{rmk}
By writing the action in terms of the Laplacian + curvature form in \eqnref{eq:Laplacian}, we have assumed to study the RG flow of this particular conformal field theory (CFT). However, RG flow can be more generally defined for any field theory with any action to start with, not necessarily near a conformal fixed point. See Sec.\,\ref{sec:fixed point} for more discussions of RG flow around general fixed points.
\end{rmk}

\subsection{Deformation family of Lagrangian action integrals} Let us denote$$\mathcal{S}(f, \phi_{0}):=\int_{C}\mathcal{L}(f, \phi_{i}, i=1,\cdots, n)=\int_{C} \lambda^{-1} \mu_{mn}(x)\partial_{m} f_{i}(x)\partial_{n} f_{j}(x)g^{ij}_{0}(f(x)),$$where $\phi_{0}:=\lambda^{-1}(g_{0},0)$ is a field associated to the fixed choice of $g_{0}$. One interesting case of study is to identify the moduli space (the geometric space representing the family) of smooth maps $f: C\to X$ which minimize the action integral $\mathcal{S}(f, \phi_{0})$ for fixed choice of metric $g_{0}$ over $X$. These are often identified with vacuum states of the underlying governing physical theory for our system of particles. A rather more interesting question is whether the vacuum states of the underlying theory are stable with respect to infinitesimal deformations of the geometry of $C$ and $X$ respectively, specially in quantum physics where fields and geometry of space undergo algebraic or analytic fluctuations. This question could be rigorously studied via inducing deformations of the fields involved in our physical theory, that is $$\phi_{0}\to \phi_{0}+\partial \phi_{0}=\lambda^{-1}(g_{0}+h, 0+\lambda \rho),$$where the function $h\in C^{\infty}(X, T^{\vee}X^{\otimes 2})$ is a seymmetric bilinear smooth differential form on $X$ and $\rho\in C^{\infty}(X, \mathbb{K})$ is a smooth function on $X$. Introducing these deformation parameters, one can study the set of extremizing maps $f: C\to X$ of the action integral $\mathcal{S}(f, \phi)$, that is smooth harmonic maps minimizing $\mathcal{S}(f, \phi)$, where $\mathcal{S}(f, \phi)$ is obtained as a local deformation around $\mathcal{S}(f, \phi_{0})$ induced by deforming the geometry of $C, X$. Let us consider a generalized deformed Lagrangian action

\begin{align}
&\mathcal{S}(f, \phi)=\mathcal{S}(f, \phi_{0})+\int_{C} \lambda^{-1} \mu_{mn}(x)\partial_{m} f_{i}(x)\partial_{n} f_{j}(x)g_{ij}(f(x))d\nu_{C}\notag\\
&+\lambda^{-1}\int_{C} \lambda \rho(f)\mathcal{K}d\nu_{C}+\lambda^{-1}\int_{C}\Gamma(f)d\nu_{C}+\lambda^{-1}\int_{C}f^{*}\omega d\nu_{C}
\end{align}
where as before $h\in C^{\infty}(X, T^{\vee}X^{\otimes 2})$, $\Gamma\in C^{\infty}(X, \mathbb{K})$, and $\omega\in C^{\infty}(X, \wedge^{2}T^{\vee}X)$ an antisymmetric bilinear form are all regarded as infinitesimal induced deformation parameters. Note that here the deformation parameters $\phi_{1}:=\lambda^{-1}h$, $\phi_{2}=\lambda^{-1} (\lambda \rho)$, $\phi_{3}:=\lambda^{-1}U$ and $\phi_{4}:= \lambda^{-1}\omega$ may, roughly speaking, be regarded as local coordinates in the space of deformations of $\mathcal{S}(f, \phi_{0})$. Hence we can rewrite one such deformation in terms of the other as an extension 
\begin{equation}\label{local-coord}
\mathcal{S}(f, \phi)= \mathcal{S}(f, \phi_{0})+\sum_{i\geq 1}\int_{C}O_{i}(f, \phi_{i}).
\end{equation}

Moreover, it must be noted that depending on the underlying physical theory, one may consider situations where $\mathcal{S}(f, \phi_{0})$ is required to be invariant under conformal transformations $(C, \mu_{mn})\to (C, e^{-\psi}\mu_{mn})$, in which case, shall one be interested to preserve the conformal invariance of the deformed Lagrangian action $\mathcal{S}(f, \phi)$, one requires that the deformation fields $\rho$ and $U$ vanish, as they break the conformal symmetry, however the deformations $h, \omega$ can be non-vanishing, as their associated integrals are preserved under conformal group action on $C$. 
\subsection{Moduli functors associated to deforming fields and maps  simultaneously} We mimic the approach of algebraic geometers for constructing our moduli spaces. Consider the following situation. Let $\mathcal{T}\to \text{Spec}{\mathbb{K}}$ be a finite type parametrizing scheme. The notation means that $\mathcal{T}$ is a space (known as parametrizing scheme in algebraic geometry terms) constructed over field of numbers $\mathbb{K}$ that is topologically compact. Let $\mathfrak{M}\text{ap}(C, X): \mathcal{S}\text{ch}/\mathbb{K}\to \mathcal{A}\text{b}/ \mathbb{K}$ be defined as a two-category (i.e. a category which contains objects, their morphisms, and their morphisms of morphisms , also known as 2-morphisms), such that the category is fibered over a base category of finite type (parametrizing) schemes over $\mathbb{K}$. The objective of such functor is to produce families of maps from $C$ to $X$ parametrized by schemes such as $\mathcal{T}$. To state the latter functionality of $\mathfrak{M}\text{ap}(C, X)$ in more mathematical formal terms, we say that the groupoid sections of $\mathfrak{M}\text{ap}(C, X)$ over any $\mathcal{T}$ are given by the sheaf of Abelian groups of $\mathcal{T}$-families of smooth maps from $C$ to $X$, that is the groupoid sections of our moduli functor are given by families of maps 
\begin{equation}\label{T-family}
\mathfrak{M}\text{ap}(C, X)(\mathcal{T})\cong \{\tilde{f}: C_{\mathcal{T}}:=\mathcal{T} \times_{\mathbb{K}} C\to X\}
\end{equation} such that for any $t\in \mathcal{T}$ the $t$-fibers of the family $\tilde{f}\mid_{t}\cong \{ f_{t}: C\to X\}$ are given by smooth continuous maps from domain Riemann surface $C$ to $X$. Roughly speaking, the functor $\mathfrak{M}\text{ap}(C, X)$ provides us with a platform to parametrize the smooth maps from $C$ to $X$ in a systematic way over any chosen parametrizing scheme. For instance, given any $\mathcal{T}:=\text{Spec}(\mathbb{K})$, geometric reduced point, the groupoid sections of $\mathfrak{M}\text{ap}(C, X)(\mathcal{T})$ are given by single maps $f:C\to X$. Similarly, the fibers of $\mathfrak{M}\text{ap}(C, X)$ over a line, $L$ (which as a geometric scheme belongs to our category, $\mathcal{S}\text{ch}/\mathbb{K}$, of schemes of $\mathbb{K}$) provides a one dimensional family of maps $f_{L}: C_{L}\to X$, and the fibers of $\mathfrak{M}\text{ap}(C, X)$ over a surface provides a two dimensional family of maps, etc.\\

Now as the geometric structure of $C,X$, and hence $f$ undergo deformations in our theory, similar to Feynman path integration formalism, we compute the vaccum states of the theory, by taking a stochastic average over all admissible weighted morphisms $f: C\to X$ which satisfy smoothness property. In doing so, we further allow certain induced correlation fields, defined in our theory, induced by evaluating the map $f$ at a finite number of smooth distinct marked points $p_{1}, \cdots, p_{l} \in C$. Moreover, we use the Lagrangian action integral constructed in previous section as a weight function associated to each single map $f:C\to X$. Doing so, we obtain an integral over the space parametrizing tuples $(f: C\to X, p_{1}, \cdots, p_{l})$, where $p_{i}, i=1, \cdots ,l$ are distinct smooth marked points on $C$

\begin{equation}\label{correlation}
Z[C, X, p_{1}, \cdots, p_{n}, \phi]:= \frac{1}{Z_{0}} \int_{\mathfrak{M}\text{ap}(C, p_{1}, \cdots, p_{l},X)} D_{\phi}[f](f(p_{1}, \cdots, f(p_{l}))) e^{-\mathcal{S}(f, \phi)}.
\end{equation}
Here $D_{\phi}(f)$ is a measure over $\mathfrak{M}\text{ap}(C, p_{1}, \cdots, p_{l},X)$. Note that by construction $\mathcal{S}(f, \phi)$ is regarded as a deformation of $\mathcal{S}(f, \phi_{0})$, hence following the construction in \eqref{local-coord}, one may rewrite correlation function \eqref{correlation} in terms of $\mathcal{S}(f, \phi_{0})$ as follows
\begin{align}\label{correlator}
&Z[C, X, p_{1}, \cdots, p_{n}, \phi]=\notag\\
&\frac{1}{Z_{0}} \int_{\mathfrak{M}\text{ap}(C, p_{1}, \cdots, p_{l},X)} D_{\phi}[f](f(p_{1}, \cdots, f(p_{l}))) e^{-\mathcal{S}(f, \phi_{0})} \prod_{i\geq 1}\int_{C} O_{i}(f, \phi_{i})
\end{align}
\section{Renormalization semi-group flow}
The construction of the renormalization semi-group flow is based on the fact that, in order to make the above integrals well-defined, one  may merely consider certain controllable deformation regimes for the fields $\phi_{i}$, that is; one would like to consider a family of the fields $\phi_{i}(\mathcal{T})$, where the scheme $\mathcal{T}$ is the parametrizing scheme, used in \eqref{T-family} governing the geometric deformations of maps $f_{\mathcal{T}}: C_{\mathcal{T}} \to X$ induced by perturbation of geometric structures of $C$ and $X$. The idea is to consider an infinitesimal deformation flow, called renormalization semi-group flow (as it turns out that our construction in this example only provides a semi-group rather than  a group), over the moduli space of maps and field deformations, that is, to consider a morphism

\begin{align}\label{TT-family}
&\mathcal{RG}_{\mathcal{T}}:=\mathfrak{M}\text{ap}(C, X)(\mathcal{T})\times H^{*}(X, \mathbb{K})^{\otimes n} \to \mathfrak{M}\text{ap}(C, X)(\mathcal{T})\times H^{*}(X, \mathbb{K})^{\otimes n}\notag\\
&\,\,\,\,\,\,\,\,\,\,\,\,\,\,\,\,\,\,\,\,\,\,\,\,\,\,\,\,\,\,\,\,\,\,\,\,\,\,\,\,\,\,\,\,\,\,\,\,\,\,\,\,\,\,\,\,\,\,\,\,\,\, (f, \phi_{1}, \cdots, \phi_{n}) \to (f_{\mathcal{T}}, \phi_{1,\mathcal{T}},\cdots, \phi_{n,\mathcal{T}})
\end{align}
which has a lift to a morphism on moduli space of action integrals
\begin{align}
&\overline{\mathcal{RG}}_{\mathcal{T}}:=\mathcal{A}\text{ct}(C,X)\cong \left(\mathfrak{M}\text{ap}(C, X)(\mathcal{T})\times H^{*}(X, \mathbb{K})^{\otimes n}\right)^{\vee}\notag\\
&\to \mathcal{A}\text{ct}(C,X)\cong \left(\mathfrak{M}\text{ap}(C, X)(\mathcal{T})\times H^{*}(X, \mathbb{K})^{\otimes n}\right)^{\vee}\notag\\
\end{align}
taking $\mathcal{S}(f, \phi_{0})$ to $\mathcal{S}(f_{\mathcal{T}}, \phi_{\mathcal{T}})$, which satisfies the semi-group property.
\begin{rmk}
We remark again that we are considering, generally speaking, our fields $\phi_{i}, i=1,\cdots, n$ as living in our field algebra, that is the vector space $H^{*}(X, \mathbb{K})^{\otimes n}$ generated by differentially graded forms on $X$. Moreover, the action integrals are regarded as morphisms from $\mathfrak{M}\text{ap}(C, X)(\mathcal{T})\times H^{*}(X, \mathbb{K})$ to the underlying ground field $\mathbb{K}$, and hence, realized as the dual space $\left(\mathfrak{M}\text{ap}(C, X)(\mathcal{T})\times H^{*}(X, \mathbb{K})^{\otimes n}\right)^{\vee}$.
\end{rmk}
 We now elaborate further on renormalization flow. In order to define it we need to formulate a deformation process, applied to geometry of $C, X$, then compute the induced deformations of associated fields $\phi_{i}$ and $f$ with support on deformed $X$ as shown in Equation \eqref{TT-family}. Note that the functorial construction of the moduli space of maps allows us to perform this task in a rigorous algebraic manner. Take a scheme $\mathcal{T}$ (naively speaking schemes have as their skeleton, the geometrical spaces however, they come further equipped with extra topological or algebraic properties). As we noted above, the fibers of the moduli functor $\mathfrak{M}\text{ap}(C, X)$ over $\mathcal{T}$ (i.e. $\mathfrak{M}\text{ap}(C, X)(\mathcal{T})$) provide us with a $\mathcal{T}$-family of maps from $C\to X$ as in \eqref{T-family}. Now choose an algebraic deformation (a perturbation) of $\mathcal{T}$ and denote it by $\mathcal{T}'$. Then the fibers $\mathfrak{M}\text{ap}(C, X)(\mathcal{T}')$ provide a $\mathcal{T}'$-family $$\{\tilde{f}': C_{\mathcal{T}'}:=\mathcal{T}' \times_{\mathbb{K}} C\to X\},$$realized as a deformation of the former $\mathcal{T}$-family maps from $C$ to $X$.

 One way of constructing such algrbraic deformation is to construct $\mathcal{T}'$ as a \textit{nilpotent thickening} of $\mathcal{T}$. We elaborate on this notion, using the language of ideals over the ring of polynomial functions. 
 
 Take the polynomial ring $\mathbb{C}[x_{1}, \cdots, x_{n}]$. In classical algebraic geometry, the set of prime ideals generated by different expressions involving the variables $x_{1}, \cdots, x_{n}$ makes a space,  isomorphic to the ``\textit{affine}" space $\mathbb{C}^{n}$. Now in order to obtain more interesting spaces, one may consider an ideal, say as an example $\mathcal{I}=(x_{1}x_{2}-x_{3}^2)$, and consider the quotient ring $\mathbb{C}[x_{1}, \cdots, x_{n}]/ \mathcal{I}$. This expression means that all polynomials generated by the expression $x_{1}x_{2}-x_{3}^2$ vanish on this quotient ring. Now the set of prime ideals $p\subset \mathbb{C}[x_{1}, \cdots, x_{n}]/ \mathcal{I} $ provides us with set of geometric points of the algebraic space (algebraic variety) given as the solution set to the polynomial equation $x_{1}x_{2}-x_{3}^2=0$. Let us denote this algebraic variety as $\mathcal{T}$. In order to obtain a nilpotent thickening of $\mathcal{T}$ one can simply construct the quotient ring $\mathbb{C}[x_{1}, \cdots, x_{n}]/ \mathcal{I}^{l}$ for some $l$. The set of prime ideals in the latter provides one with the set of geometric points of the variety obtained as the solution set to $(x_{1}x_{2}-x_{3}^2)^{l}=0$, call the latter space as $\mathcal{T}'$. Due to the natural inclusion of ideals $\mathcal{I}^{l}\subset \mathcal{I}$, one can immediately obtain a natural inclusion of $\mathcal{T}\hookrightarrow \mathcal{T}'$. This deformation is called a nilpotent extension of $\mathcal{T}$ of order $l$. Given a deformation as such nilpotent extension, $\iota_{TT'}:T\hookrightarrow T'$, as we elaborated earlier, the renormalization flow must satisfy the property that$$\iota_{TT'}^{*}\overline{\mathcal{RG}}_{\mathcal{T}}\left(\mathcal{S}(f, \phi_{0})\right)=\mathcal{S}(\iota_{TT'}^{*}\mathcal{RG}_{\mathcal{T'}}(f, \phi_{0})).$$Since the action of the RG flow is realized as a pullback in our construction, one is able to define its induced action on the correlation function defined in \eqref{correlator} as follows

\begin{align}
&\overline{\mathcal{RG}}_{\mathcal{T}}(Z[C, X, p_{1}, \cdots, p_{n}, \phi])= \frac{1}{Z_{0}} \int_{\mathcal{RG}_{\mathcal{T}}(\mathfrak{M}\text{ap}(C, p_{1}, \cdots, p_{l},X))} D_{\phi}[f](f(p_{1}, \cdots, f(p_{l}))) e^{-\mathcal{S}(f, \phi)}\notag\\
&=\frac{1}{Z_{0}} \int_{\mathfrak{M}\text{ap}(C, p_{1}, \cdots, p_{l},X)} \mathcal{RG}_{\mathcal{T}}^{*}(D_{\phi}[f](f(p_{1}, \cdots, f(p_{l})))) e^{-\overline{\mathcal{RG}}_{\mathcal{T}}^{*}(\mathcal{S}(f, \phi))} \notag\\
\end{align}
Let us work out a concrete example. 
\begin{exam}
For simplicity, let us assume that $\mathbb{K}$ is given as a field of characteristic zero, such as $\mathbb{C}$, the field of complex numbers. Consider the case where $\mathcal{T}:= \text{Spec}(\mathbb{K}[x_{1}, x_{2}, \cdots, x_{n}]/ (x_{2}, \cdots, x_{n}))\cong \mathbb{A}^{1}$ is given by taking the Zariski spectrum of the affine line in the direction $x_{1}$, given by ideal $\mathcal{I}=(x_{2}, \cdots, x_{n})$ over $\mathbb{K}$. Locally, after choosing a coordinate chart $(x_{1}, \cdots, x_{n})$, the set geometric points in $\mathcal{T}$ is the set of points on the $x_{1}$ axis in $\mathbb{C}^{n}$. Now we introduce an infinitesimal deformation of $\mathcal{T}\hookrightarrow \mathcal{T}'$, induced by a nilpotent extension of order 2, by taking $\mathcal{T}':=\text{Spec}(\mathbb{K}[x_{1}, \cdots, x_{n}]/\mathcal{I}^{2})$. There exists a canonical short exact sequence $$0\to \mathcal{I}/\mathcal{I}^{2}\to \mathbb{K}[x_{1}, \cdots, x_{n}]/\mathcal{I}^{2} \to \mathbb{K}[x_{1}, \cdots, x_{n}]/\mathcal{I}\to 0$$whose kernel is governed by the conormal sheaf (which in here is identified by sheaf of differential one forms on $\mathcal{T}$, that is $\Omega_{\mathcal{T}}$). This roughly speaking realizes the second order nilpotent thickening of $\mathcal{T}$, as the cotangent bundle, $\Omega_{\mathcal{T}}$, of $\mathcal{T}$. We would like to deform the correlation function \eqref{correlator} in the direction of fibers of $\Omega_{\mathcal{T}}$. This amounts to setting $\mathcal{RG}_{\mathcal{T}}$ as the differential operator which deforms the fields in direction of fibers of cotangent bundle of $\mathcal{T}$, that is, RG flow acts on the fields as a map $\phi\to \phi+d\phi$ and hence its induced action on the action integral is given by
\tikzset{every picture/.style={line width=0.75pt}} 

\begin{tikzpicture}[x=0.75pt,y=0.75pt,yscale=-.5,xscale=.5]

\draw    (250,171) -- (252,334.5) ;
\draw    (131,448.5) -- (252,334.5) ;
\draw    (373,218.5) -- (252,334.5) ;
\draw    (252,334.5) -- (517,409.5) ;
\draw    (131,379.5) -- (370,152.5) ;
\draw    (131,379.5) -- (131,446.5) ;
\draw [shift={(131,448.5)}, rotate = 270] [color={rgb, 255:red, 0; green, 0; blue, 0 }  ][line width=0.75]    (10.93,-3.29) .. controls (6.95,-1.4) and (3.31,-0.3) .. (0,0) .. controls (3.31,0.3) and (6.95,1.4) .. (10.93,3.29)   ;
\draw    (155,357.5) -- (155,424.5) ;
\draw [shift={(155,426.5)}, rotate = 270] [color={rgb, 255:red, 0; green, 0; blue, 0 }  ][line width=0.75]    (10.93,-3.29) .. controls (6.95,-1.4) and (3.31,-0.3) .. (0,0) .. controls (3.31,0.3) and (6.95,1.4) .. (10.93,3.29)   ;
\draw    (177,334.5) -- (177,401.5) ;
\draw [shift={(177,403.5)}, rotate = 270] [color={rgb, 255:red, 0; green, 0; blue, 0 }  ][line width=0.75]    (10.93,-3.29) .. controls (6.95,-1.4) and (3.31,-0.3) .. (0,0) .. controls (3.31,0.3) and (6.95,1.4) .. (10.93,3.29)   ;
\draw    (212,302.5) -- (212,369.5) ;
\draw [shift={(212,371.5)}, rotate = 270] [color={rgb, 255:red, 0; green, 0; blue, 0 }  ][line width=0.75]    (10.93,-3.29) .. controls (6.95,-1.4) and (3.31,-0.3) .. (0,0) .. controls (3.31,0.3) and (6.95,1.4) .. (10.93,3.29)   ;
\draw    (234,281.5) -- (234,348.5) ;
\draw [shift={(234,350.5)}, rotate = 270] [color={rgb, 255:red, 0; green, 0; blue, 0 }  ][line width=0.75]    (10.93,-3.29) .. controls (6.95,-1.4) and (3.31,-0.3) .. (0,0) .. controls (3.31,0.3) and (6.95,1.4) .. (10.93,3.29)   ;
\draw    (271,247.5) -- (271,314.5) ;
\draw [shift={(271,316.5)}, rotate = 270] [color={rgb, 255:red, 0; green, 0; blue, 0 }  ][line width=0.75]    (10.93,-3.29) .. controls (6.95,-1.4) and (3.31,-0.3) .. (0,0) .. controls (3.31,0.3) and (6.95,1.4) .. (10.93,3.29)   ;
\draw    (295,223.5) -- (295,290.5) ;
\draw [shift={(295,292.5)}, rotate = 270] [color={rgb, 255:red, 0; green, 0; blue, 0 }  ][line width=0.75]    (10.93,-3.29) .. controls (6.95,-1.4) and (3.31,-0.3) .. (0,0) .. controls (3.31,0.3) and (6.95,1.4) .. (10.93,3.29)   ;
\draw    (317.5,201.5) -- (317.5,268.5) ;
\draw [shift={(317.5,270.5)}, rotate = 270] [color={rgb, 255:red, 0; green, 0; blue, 0 }  ][line width=0.75]    (10.93,-3.29) .. controls (6.95,-1.4) and (3.31,-0.3) .. (0,0) .. controls (3.31,0.3) and (6.95,1.4) .. (10.93,3.29)   ;
\draw    (339,182.5) -- (339,249.5) ;
\draw [shift={(339,251.5)}, rotate = 270] [color={rgb, 255:red, 0; green, 0; blue, 0 }  ][line width=0.75]    (10.93,-3.29) .. controls (6.95,-1.4) and (3.31,-0.3) .. (0,0) .. controls (3.31,0.3) and (6.95,1.4) .. (10.93,3.29)   ;
\draw    (196,318.5) -- (196,385.5) ;
\draw [shift={(196,387.5)}, rotate = 270] [color={rgb, 255:red, 0; green, 0; blue, 0 }  ][line width=0.75]    (10.93,-3.29) .. controls (6.95,-1.4) and (3.31,-0.3) .. (0,0) .. controls (3.31,0.3) and (6.95,1.4) .. (10.93,3.29)   ;
\draw    (405,177.5) -- (376.98,181.24) ;
\draw [shift={(375,181.5)}, rotate = 352.40999999999997] [color={rgb, 255:red, 0; green, 0; blue, 0 }  ][line width=0.75]    (10.93,-3.29) .. controls (6.95,-1.4) and (3.31,-0.3) .. (0,0) .. controls (3.31,0.3) and (6.95,1.4) .. (10.93,3.29)   ;

\draw (277,445) node [anchor=north west][inner sep=0.75pt]   [align=left] {$\displaystyle {\textstyle \cdots }$};
\draw (149,434.4) node [anchor=north west][inner sep=0.75pt]    {$x_{1}$};
\draw (508,413.4) node [anchor=north west][inner sep=0.75pt]    {$x_{n}$};
\draw (227,138.4) node [anchor=north west][inner sep=0.75pt]    {$x_{2}$};
\draw (413.88,160.55) node [anchor=north west][inner sep=0.75pt]  [font=\Large,rotate=-359.08]  {$\mathcal{T} ':\ Tot\left( \Omega _{\mathcal{T}}\rightarrow \mathcal{T}\right)$};
\draw (352.88,243.55) node [anchor=north west][inner sep=0.75pt]  [font=\Large,rotate=-359.08]  {$\mathcal{T}$};
\draw (70,110) node   [align=left] {\begin{minipage}[lt]{68pt}\setlength\topsep{0pt}
{\large{$\mathbb{A}^{n}$}}
\end{minipage}};

\end{tikzpicture}

\begin{align}
&\overline{\mathcal{RG}}_{\mathcal{T}}(Z[C, X, p_{1}, \cdots, p_{n}, \phi])= \frac{1}{Z_{0}} \int_{\mathcal{RG}_{\mathcal{T}}(\mathfrak{M}\text{ap}(C, p_{1}, \cdots, p_{l},X))} D_{\phi}[f](f(p_{1}, \cdots, f(p_{l}))) e^{-\mathcal{S}(f, \phi)}\notag\\
&=\frac{1}{Z_{0}} \int_{\mathfrak{M}\text{ap}(C, p_{1}, \cdots, p_{l},X)} \mathcal{RG}_{\mathcal{T}}^{*}(D_{\phi}[f](f(p_{1}, \cdots, f(p_{l})))) e^{-\overline{\mathcal{RG}}_{\mathcal{T}}^{*}(\mathcal{S}(f, \phi))} \notag\\
&\frac{1}{Z_{0}} \int_{\mathfrak{M}\text{ap}(C, p_{1}, \cdots, p_{l},X)} (D_{\phi+d\phi}[f](f(p_{1}, \cdots, f(p_{l})))) e^{-(\mathcal{S}(f, \phi+d\phi))}\notag\\
&=-\frac{1}{Z_{0}} \int_{\mathfrak{M}\text{ap}(C, p_{1}, \cdots, p_{l},X)} \frac{\partial \phi(x_{n})}{\partial x_{n}} \frac{\partial}{\partial \phi} \left( D_{\phi}[f](f(p_{1}, \cdots, f(p_{l}))) e^{-\mathcal{S}(f, \phi)}\right)\notag\\
\end{align} 
Therefore, viewing the RG flow as a differential operator acting on the action integral, $Z$, and rewriting the variation of $Z$, induced by nilpotent deformation of $\mathcal{T}$, in terms of $Z$ itself, we obtain a differential equation governing the change in $Z$, that is 
\begin{equation}
\left[ \overline{\mathcal{RG}}_{\mathcal{T}}+\frac{\partial \phi(x_{n})}{\partial x_{n}} \frac{\partial}{\partial \phi}\right]Z[C, X, p_{1}, \cdots, p_{n}, \phi]=0
\end{equation}
We will come back to the discrete version of the differential equation above, when discussing the construction of RG flow in AI.
\end{exam}

\section{From Conventional RG to Machine-Learning RG}

The above section formulates the mathematical foundation for conventional RG. However, there are several aspects that should be upgraded before the idea of RG can find useful applications in machine learning. The main differences between the conventional RG and the machine-learning RG are summarized in \tabref{tab:RG_compare}, and discussed as follows.

\begin{table}[htp]
\caption{Differences between conventional RG and machine-learning RG proposed in Ref.\,\cite{Li2018Neural,Hu2020Machine}}
\begin{center}
\begin{tabular}{ccc}
& Conventional RG & Machine-Learning RG\\
\hline
base space & smooth manifold & discrete lattice\\
RG flow & continuous & discrete \\
RG equation & differential equation & recurrence equation \\
RG fixed point & conformal & general \\
RG scheme & human-specified (fixed) & machine-designed (learnable) \\ 
data driven & no & yes \\
algebraic structure & semigroup & group \\
invertibility & non-invertible & invertible \\
holographic bulk & not available & available \\
hyperbolic geometry & not defined & emergent
\end{tabular}
\end{center}
\label{tab:RG_compare}
\end{table}%

\subsection{Continuous v.s. Discrete}

The conventional RG in the quantum field theory typically assumes that the field is defined on a smooth base manifold. However, this assumption is typically not the case for machine learning applications. For example, images are defined on discrete pixels, and texts are defined on discrete words. The discrete nature of most datasets in machine learning requires us to generalize the base manifold from continuous space to discrete lattice.

The discretization of the base manifold also forces the RG flow to be discrete, because it is no longer possible to perform infinitesimal dilation on a discrete lattice. Therefore, instead of writing down a differential equation to describe the continuous RG flow, the discrete RG flow should be described by a \emph{recurrent} equation. However, in the continuum limit (when the lattice spacing approaches to zero), the recurrent equation should converge to the differential equation, which will be shown in Sec.\,\ref{sec:conventional RG}.

\subsection{Semigroup v.s. Group} The conventional RG keeps decimating information in each step of coarse-graining. As a result, the conventional RG is not invertible and only forms a \emph{semigroup} instead of a \emph{group}, despite of its inaccurate name of renormalization ``group''. Recent development in physics\cite{Qi2013Exact} reveals that the RG flow can actually be viewed as a holographic mapping, which is invertible. This not only makes a profound connection from RG to quantum gravity, but also promotes RG to a group. 

The conventional RG studies how perturbations of the action (or deformations of the field configuration) gets renormalized at larger and larger scales. The invertible RG has a completely different mindset: it aims to answer how the correlated field configurations on the holographic boundary can be disentangled to uncorrelated noises in the holographic bulk, or how the strongly-coupled quantum field theory  on the holographic boundary can be reformulated as the weakly-coupled dual gravitational theory in the holographic bulk. By establishing the holographic mapping, any deformation of the field configuration on the holographic boundary can be translated into an excitation in the holographic bulk and analyzed more conveniently. Therefore invertible RG is a more powerful paradigm of RG. Nevertheless, it can always fall back to the conventional RG by a forgetful map that forgets about the holographic bulk degrees of freedom.

\subsection{Human v.s. Machine} The conventional RG scheme is designed by human. Due to the limitation of human intelligence, the conventional RG always assumes that the action must take a fixed form with specific types of terms, and the RG flow only change the coefficients of these terms, such that the action can only flow within a predefined moduli space. Although the moduli space allows us to parameterize the action conveniently, it also restricts our imagination. A more general RG flow can go beyond the moduli space, as new terms can be generated under RG and even the field content can change under RG (microscopic and macroscopic descriptions of a system can be fundamentally different as advocated by the emergence principle). However, such a general RG scheme is not analytically tractable by human. It is not even clear how to design the RG scheme if the form of the action and the field content are all unknown. Thus it becomes desirable to introduce artificial intelligence to learn the optimal RG scheme automatically from the big data of field configurations generated by a field theory. By learning to generate similar field configurations from independent random noise in the holographic bulk, the machine will create the optimal holographic mapping, which also specifies the optimal (invertible) RG scheme.

\subsection{Conformal v.s. General Fixed Point}\label{sec:fixed point} Conventional RG typically assume a conformal fixed point to start with. Given the conformal symmetry at the fixed point, the RG transformation is always taken to be the dilation operator in the conformal group, which corresponds to the rescaling of spacetime and fields together. Given the RG transformation, one can study how a perturbation (or deformation) of the field evolves under dilation. If the perturbation grows stronger/weaker at larger scales, then the perturbation is said to be \emph{relevant}/\emph{irrelevant} (with respect to the conformal fixed point). More quantitatively, the conformal dimensions can be defined as the eigenvalues of the dilation generator, such that relevant/irrelevant fields are simply distinguished by their positive/negative conformal dimensions. Intuitively, relevant fields are low-energy/slow-varying modes to be kept under coarse-graining, and irrelevant fields are high-energy/fast-varying modes to be decimated (or integrated out).

However, the more general machine-learning RG do not assume a conformal fixed point, because the real-world data (like images or texts) may not be scale-invariant and hence not respecting the conformal symmetry. Therefore, the dilation operator is not well-defined, and one can not prescribe an explicit RG scheme from the beginning. The RG scheme has to be learned from data using a data-driven approach. In fact, the real-world data is more likely to be closer to Gaussian fixed points. So even if one learns the RG scheme, it is not immediately clear whether the RG transformation can be used to infer the conformal dimension, as the data could be far from any conformal fixed point.

\subsection{Relevant vs. Irrelevant}

Therefore, the traditional idea of calculating scaling dimension as eigenvalues of the dilation generator no longer works in more general RG approaches. We need a different way to define what is relevant and what is irrelevant. Ref.\,\cite{Hu2020Machine} proposes an elegant and universal definition of irrelevant degrees of freedom using holographic duality and information theory. The key idea is that irrelevant fields are those degrees of freedoms that should be decimated under coarse-graining, so they should appear to us as random noise (i.e.~independent/uncorrelated random variables). Since the irrelevant fields are actually the holographic bulk field under the holographic duality, the above idea can also be rephrased to a statement that holographic bulk fields are almost uncorrelated. The goal of machine-learning RG is to learn the RG transformation that automatically identify and separate such irrelevant degrees of freedom in a field theory. We will explain this approach in more details in Sec.\,\ref{sec:irrelevance}, after introducing the concrete construction of the machine-learning RG algorithm. 

For now, we would like to comment that the information theoretic definition of the irrelevant field is consistent with the conformal dimension definition in the conformal limit. Because a \emph{negative} conformal dimension in the conformal field theory (CFT) indicates that the field correlation will \emph{decay} exponentially in the dual anti-de Sitter (AdS) holographic bulk, which is equivalent to the statement that the holographic bulk field are \emph{short-range correlated}, which look like independent random noises beyond a finite correlation length, and are therefore \emph{irrelevant} in the information theoretic sense. 

\section{Machine-Learning RG via Flow-Based Generative Models}

\subsection{Sequential Data and Quantum Field on One-Dimensional Lattice}\label{1dlattice}

The idea of renormalization group can be used to construct novel generative models for unsupervised learning. The discussion will mainly focus on sequential data, although generalizations to images and graphs are possible. A sequence is an ordered set of objects $a=(a_1,a_2,\cdots)$, where each object $a_i\in A$ is taken from an object set $A$ (also known as the vocabulary). In machine learning, each object $a_i$ is usually embedded as a vector $\phi_i$ in a finite-dimensional vector space $\dsR^n$ (assuming the dimension to be $n$). Denote the embedding map as $E:A\to\dsR^n$, the sequence can be represented as a ordered set of vectors $\phi=(\phi_1,\phi_2,\cdots)$, where $\phi_i = E(a_i)\in\dsR^n$. 

One can also view $\phi_i$ as a quantum field on one-dimensional discrete lattice, as described by the mapping $\phi:I\to\dsR^n$, where $I\subset\dsN$ denotes the index set (equipped with an ordering). Each index $i\in I$ labels an object (or its vector embeding) in the sequence and the set $I$ describes the one-dimensional lattice. The size (cardinality) $|I|$ of the index set corresponds to the length of the sequence. Let $\mathsf{Map}(I, \dsR^n):=\{\phi: I\to \dsR^n\}$ be the associated space of all maps from the index set $I$ to the vector space $\dsR^n$. The objective of unsupervised machine learning is to model the probability measure $p(\phi)\scD\phi$ given the dataset of sequences.

\subsection{Conventional Renormalization froms a Semigroup}

The conventional notion of renormalization group transformation $\scR:\mathsf{Map}(I,\dsR^n)\to\mathsf{Map}(I',\dsR^n)$ corresponds to a coarse-graining map that extract the relevant (coarse-grained) field $\phi'=\scR(\phi)$ from the original (fine-grained) field $\phi$ and discard the remaining (irrelevant) field degrees of freedom. The renormalization transformation always reduces the degrees of freedom, therefore the index set will become smaller $|I'|\leq |I|$ under the renormalization transformation. Because of the information loss, it is no-longer possible to recover the original field configuration $\phi$ from the coarse-grained configuration $\phi'$. Therefore the renormalization transformation $\scR$ is not invertible, and only forms a \emph{semigroup}.

\subsection{Invertible Renormalization forms a Group}

The key idea to make the renormalization transformation invertible is to keep  the irrelevant field $\zeta'$ together with the relevant field $\phi'$ as the joint output of the renormalization transformation. Intuitively, the relevant/irrelevant fields are the low-/high-energy modes in the field configuration. What the renormalization transformation does is to separate the irrelevant field $\zeta'$ and the relevant $\phi'$ field given the original field $\phi$ as input. The criterion to separate irrelevant fields will be elaborated in Sec.\,\ref{sec:irrelevance}. 

Invertible renormalization was first proposed under the name of exact holographic mapping (EHM) \cite{Qi2013Exact}, which further leads to applications in flow-base generative models for unsupervised machine learning \cite{Li2018Neural,Hu2020Machine,Hu2020RG-Flow}. An invertible renormalization transformation is a bijective map $\hat{\scR}:\mathsf{Map}(I,\dsR^n)\to\mathsf{Map}(I', \dsR^n)\otimes\mathsf{Map}(J', \dsR^n)$, under which the original field $\phi$ splits to the relevant field $\phi'$ and the irrelevant field $\zeta'$
\begin{equation}
(\phi',\zeta') = \hat{\scR}(\phi),
\end{equation}
where $\phi'=(\cdots,\phi'_{i'},\cdots)_{i'\in I'}\in \mathsf{Map}(I', \dsR^n)$ and $\zeta'=(\cdots,\zeta'_{j'},\cdots)_{j'\in J'}\in \mathsf{Map}(J', \dsR^n)$. The bijectivity requires $|I'|+|J'|=|I|$, i.e. the numbers of relevant and irrelevant features must add up to the total number of features in the original field. The inverse renormalization transformation $\hat{\scR}^{-1}$ will also be called the generation transformation $\hat{\scG}$, denoted as 
\begin{equation}
\phi = \hat{\scG}(\phi',\zeta'):=\hat{\scR}^{-1}(\phi',\zeta').
\end{equation}
As the transformation is invertible, the renormalization group (RG) is promoted from a semigroup to a \emph{group}.

\subsection{Renormalization Group Flow} The invertible renormalization transformation enables us to define invertible renormalization group (RG) flow on both the field configuration level and the probability measure (or the action) level.

\subsubsection{RG Flow on the Field Level}

Repeating the invertible renormalization transformation, an RG flow can be defined (on the field configuration level) via the following iteration
\begin{equation}\label{eq:Rk}
(\phi^{(k)},\zeta^{(k)})=\hat{\scR}^{(k)}(\phi^{(k-1)}),
\end{equation}
where $\phi^{(k)}\in\mathsf{Map}(I^{(k)},\dsR^n)$, $\zeta^{(k)}\in\mathsf{Map}(J^{(k)},\dsR^n)$ are the relevant and irrelevant fields, and $\hat{\scR}^{(k)}:\mathsf{Map}(I^{(k-1)},\dsR^n)\to\mathsf{Map}(I^{(k)}, \dsR^n)\otimes\mathsf{Map}(J^{(k)}, \dsR^n)$ is the (bijective) renormalization transformation at the $k$-th step. The condition $|I^{(k)}|+|J^{(k)}|=|I^{(k-1)}|$ is always satisfied as a necessary condition for the bijectivity. The iteration defines a flow of quantum fields, called the \emph{renormalization flow} ($\scR$-flow):
\begin{equation}\label{eq:R flow}
\begin{tikzcd}
	{\phi\equiv\phi^{(0)}} & {\phi^{(1)}} & {\phi^{(2)}} & \cdots \\
	& {\zeta^{(1)}} & {\zeta^{(2)}} & \cdots
	\arrow["\hat{\scR}^{(1)}"', from=1-1, to=1-2]
	\arrow[from=1-1, to=2-2]
	\arrow["\hat{\scR}^{(2)}"', from=1-2, to=1-3]
	\arrow[from=1-2, to=2-3]
	\arrow["\hat{\scR}^{(3)}"', from=1-3, to=1-4]
	\arrow[from=1-3, to=2-4]
\end{tikzcd}.
\end{equation}
Along the $\scR$-flow, the field configuration will be coarse-grained progressively $\phi^{(0)}\to\phi^{(1)}\to\phi^{(2)}\to\cdots$, and the relevant degrees of freedom will be reduced (as $|I^{(0)}|\geq|I^{(1)}|\geq|I^{(2)}|\geq\cdots$). Through this process, a sequence of irrelevant fields $\zeta^{(1)},\zeta^{(2)},\cdots$ is also produced, which was discarded in the conventional renormalization approach, but kept in the invertible renormalization approach. Suppose all the relevant degrees of freedom are eliminated after $K$ steps of the renormalization transformation (i.e. $|I^{(K)}|=0$), the entire $\scR$-flow corresponds to a map that encodes the original field $\phi\equiv \phi^{(0)}$ to the collection of irrelevant fields $\zeta\equiv \{\zeta^{(k)}\}_{k=1:K}$, denoted as $\zeta=\hat{\scR}(\phi)$.

Retaining these irrelevant fields allows the RG flow to be inverted. The inverse flow is also called the \emph{generation flow} ($\scG$-flow) that reconstructs the original field, as defined by the following inverse iteration
\begin{equation}
\phi^{(k-1)}=\hat{\scG}^{(k)}(\phi^{(k)},\zeta^{(k)}):=(\hat{\scR}^{(k)})^{-1}(\phi^{(k)},\zeta^{(k)}),
\end{equation}
or given by the dual diagram of \eqnref{eq:R flow}
\begin{equation}\label{eq:G flow}
\begin{tikzcd}
	\cdots & {\phi^{(2)}} & {\phi^{(1)}} & {\phi^{(0)}\equiv\phi} \\
	\cdots & {\zeta^{(2)}} & {\zeta^{(1)}} 
	\arrow["{\hat{\scG}^{(3)}}"', from=1-1, to=1-2]
	\arrow["{\hat{\scG}^{(2)}}"', from=1-2, to=1-3]
	\arrow["{\hat{\scG}^{(1)}}"', from=1-3, to=1-4]
	\arrow[from=2-3, to=1-4]
	\arrow[from=2-2, to=1-3]
	\arrow[from=2-1, to=1-2]
\end{tikzcd}.
\end{equation}
The entire $\scG$-flow corresponds to a map that decodes the irrelevant fields $\zeta$ to the original field $\phi$, denoted as $\phi=\hat{\scG}(\zeta)$.

\subsubsection{RG Flow on the Probability Measure (Action) Level} The RG flow of field $\phi\to\zeta$ induces a flow of the associated probability distribution over $\mathsf{Map}(I,\dsR^n)$. Under the bijective map between the original field $\phi$ and the irrelevant field $\zeta$, the probability measure must remain invariant
\begin{equation}\label{eq:measure invariance}
p_{\Phi}(\phi)\scD\phi=p_{\Zeta}(\zeta)\scD\zeta.
\end{equation}
Given $\zeta=\hat{\scR}(\phi)$ and $\phi=\hat{\scG}(\zeta)$, \eqnref{eq:measure invariance} implies that the probability distributions are related by
\begin{equation}
p_{\Zeta}(\zeta)=p_{\Phi}(\phi)\Big\vert\det\frac{\partial\hat{\scR}(\phi)}{\partial\phi}\Big\vert^{-1}, \quad p_{\Phi}(\phi)=p_{\Zeta}(\zeta)\Big\vert\det\frac{\partial\hat{\scG}(\zeta)}{\partial\zeta}\Big\vert^{-1},
\end{equation}
where $|\det\partial_\phi\hat{\scR}(\phi)|$ denotes the absolute value of the Jacobian determinant of the transformation $\hat{\scR}$, and similarly for $|\det\partial_\zeta\hat{\scG}(\zeta)|$. More specifically, in each step of the transformation, the probability measure is  deformed by (along the $\scG$-flow)
\begin{equation}
p_{\Phi}^{(k-1)}(\phi^{(k-1)})=p_{\Phi}^{(k)}(\phi^{(k)}) p_{\Zeta}^{(k)}(\zeta^{(k)})\Big\vert\det\frac{\partial\hat{\scG}^{(k)}(\phi^{(k)},\zeta^{(k)})}{\partial(\phi^{(k)},\zeta^{(k)})}\Big\vert^{-1}.
\end{equation}

In quantum field theory, the field action is defined as the negative log-likelihood of the field configuration, i.e. $S_{\Phi}^{(k)}=-\log p_{\Phi}^{(k)}$ and $S_{\Zeta}^{(k)}=-\log p_{\Zeta}^{(k)}$. In terms of the field action, the transformation relates
\begin{equation}\label{eq:recursive action}
S_{\Phi}^{(k-1)}(\phi^{(k-1)})=S_{\Phi}^{(k)}(\phi^{(k)})+ S_{\Zeta}^{(k)}(\zeta^{(k)})+S^{(k)}_{\Phi\Zeta}(\phi^{(k)},\zeta^{(k)}),
\end{equation}
where the coupling action $S^{(k)}_{\Phi\Zeta}(\phi^{(k)},\zeta^{(k)})$ is defined to be the log Jacobian determinant of the $\hat{\scG}^{(k)}$ transformation,
\begin{equation}\label{eq:S_PhiZeta}
S^{(k)}_{\Phi\Zeta}(\phi^{(k)},\zeta^{(k)}):=\log\Big\vert\det\frac{\partial\hat{\scG}^{(k)}(\phi^{(k)},\zeta^{(k)})}{\partial(\phi^{(k)},\zeta^{(k)})}\Big\vert.
\end{equation}
Therefore the renormalization transformation $\hat{\scR}$ of the relevant field $\phi^{(k)}$  induces the deformation $\bar{\scG}$ of the relevant field action $S_{\Phi}^{(k)}(\phi^{(k)})$ along the generative direction
\begin{equation}
\begin{tikzcd}
	{\phi=\phi^{(0)}} & \cdots & {\phi^{(k-1)}} & {\phi^{(k)}} & \cdots & {\phi^{(K)}=\emptyset}\\
	{S_{\Phi}=S_{\Phi}^{(0)}} & \cdots & {S_{\Phi}^{(k-1)}} & {S_{\Phi}^{(k)}} & \cdots & {S_{\Phi}^{(K)}=0}
	\arrow[from=1-1, to=1-2]
	\arrow[from=1-2, to=1-3]
	\arrow["{\hat{\scR}^{(k)}}", from=1-3, to=1-4]
	\arrow[from=1-4, to=1-5]
	\arrow[from=1-5, to=1-6]
	\arrow[from=2-6, to=2-5]
	\arrow[from=2-5, to=2-4]
	\arrow["{\bar{\scG}^{(k)}}", from=2-4, to=2-3]
	\arrow[from=2-3, to=2-2]
	\arrow[from=2-2, to=2-1]
	\arrow[from=1-1, to=2-1]
	\arrow[from=1-3, to=2-3]
	\arrow[from=1-4, to=2-4]
	\arrow[from=1-6, to=2-6]
\end{tikzcd}.
\end{equation}
In this way, the renormalization flow of the action $\bar{\scR}:=\bar{\scG}^{-1}$ is defined as the pullback of the renormalization flow $\hat{\scR}$ of the field. The RG transformation is invertible on both the field and the action level, making renormalization group literally a group.

\subsection{Criterion to Separate Irrelevant Fields}\label{sec:irrelevance} What has not been explained so far is the criterion to separate relevant fields from irrelevant fields. Ref.\,\cite{Hu2020Machine} argues that the irrelevant field should look like independent random variables (or random maps), because the irrelevant fields are supposed to be discarded under the conventional RG flow, meaning that (in the ideal limit) they do not contain information and should appear like random noise. Guided by this intuition, Ref.\,\cite{Hu2020Machine} further proposes the minimal bulk mutual information (minBMI) principle as the designing principle of renormalization flow, that the optimal renormalization transformations $\{\hat{\scR}^{(k)}\}_{k=1:K}$ should be defined as the maps that minimize the mutual information among all irrelevant fields
\begin{equation}
\sum_{k,k',j,j'}I(\zeta^{(k)}_{j}:\zeta^{(k')}_{j'})=\sum_{k,k',j,j'}\int\scD\zeta\;p_{\Zeta}(\zeta)\log\frac{p_{\Zeta}(\zeta^{(k)}_{j},\zeta^{(k')}_{j'})}{p_{\Zeta}(\zeta^{(k)}_{j})p_{\Zeta}(\zeta^{(k')}_{j'})}.
\end{equation}
The minimum is achieved when the irrelevant fields are statistically independent, i.e.
\begin{equation}\label{eq:independent prior}
p_{\Zeta}(\zeta)=\prod_{k}\prod_{j\in J^{(k)}}p_{\Zeta}(\zeta^{(k)}_{j}),
\end{equation}
such that all mutual information vanishes.

The optimal solution of $\hat{\scR}$ that converges to this limit can be found using machine learning approaches, by constructing a trainable bijective map $\hat{\scG}:=\hat{\scR}^{-1}$ (as the composition of smaller bijective maps $\hat{\scG}^{(k)}$ at each RG step) to reproduce the data distribution $p_{\Phi}(\phi)$ starting from the independent prior distribution $p_{\Zeta}(\zeta)$ in \eqnref{eq:independent prior}. The related methods were developed in Refs.\,\cite{Li2018Neural,Hu2020Machine,Hu2020RG-Flow} under the name of neural-RG. A conventional choice is to take each $p_{\Zeta}(\zeta^{(k)}_{j})=(2\pi)^{-n/2}\exp(-\frac{1}{2}\Vert\zeta^{(k)}_{j}\Vert^2)$ to be the standard normal distribution (Gaussian with zero mean and unit variance), such that
\begin{equation}\label{eq:SZ}
S_{\Zeta}^{(k)}(\zeta^{(k)})=\frac{1}{2}\sum_{j\in J^{(k)}}\Vert\zeta^{(k)}_{j}\Vert^2+\text{const.}.
\end{equation}
This action describes that the irrelevant field fluctuation is massive in the holographic bulk, which is compatible with the idea of holographic duality.

\subsection{Hierarchical Structure and Hyperbolic Space}
\label{Hierarchical Structure}

As the renormalization transformation reduces the relevant degrees of freedom, the size of the relevant index set gradually reduces $|I^{(k)}|\leq |I^{(k-1)}|$. To be more concrete, we can restrict our discussion to the case where the degrees of freedom is reduced by half under each renormalization transformation, i.e. $|I^{(k)}|=|I^{(k-1)}|/2$, such that
\begin{equation}
|I^{(k)}|=2^{-k}|I^{(0)}|.
\end{equation}
Then the condition $|I^{(k)}|+|J^{(k)}|=|I^{(k-1)}|$ implies $|J^{(k)}|=2^{-k}|I^{(0)}|$. The RG flow will stop when $|I^{(K)}|<1$, which sets the total number $K$ of RG steps to be
\begin{equation}
K=\log_2|I^{(0)}|.
\end{equation}

\begin{figure}[htbp]
\begin{center}
\includegraphics[width=0.75\columnwidth]{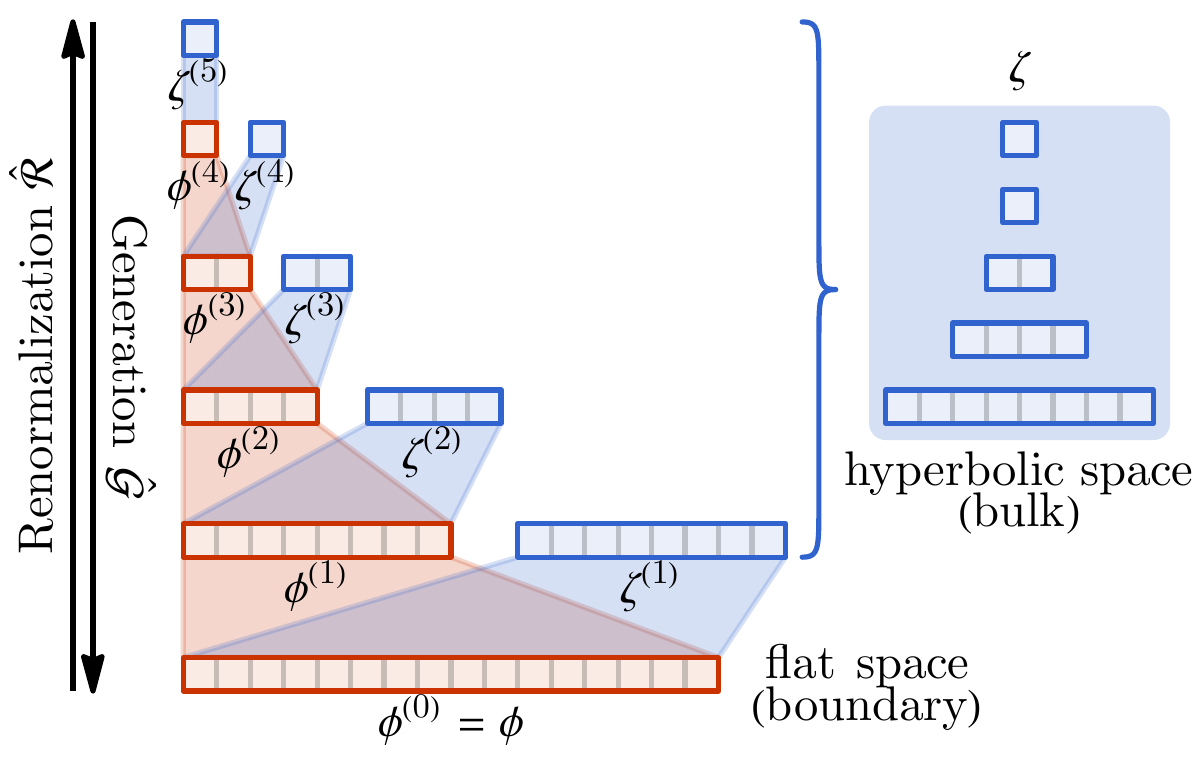}
\caption{Hierarchical structure of the RG flow. The renormalization/generation flows can be viewed as the encoding/decoding maps between the original field in the flat space (holographic boundary) and the irrelevant field in the hyperbolic space (holographic bulk).}
\label{fig:hyperbolic}
\end{center}
\end{figure}

As illustrated in \figref{fig:hyperbolic}, the hierarchical structure of the RG flow generates a ordered collection of index sets $\{J^{(k)}\}_{k=1:K}$, which can be union into a hyperbolic lattice (a discrete hyperbolic space), described by
\begin{equation}
J=\bigcup_{k=1:K} J^{(k)}.
\end{equation}
Instead of thinking the irrelevant fields as separate mappings $\zeta^{(k)}\in\mathsf{Map}(J^{(k)},\dsR^n)$, we can treat them jointly as a field $\zeta\in\mathsf{Map}(J,\dsR^n)$ defined on the hyperbolic lattice $J$. Therefore, the $\scR$-flow $\zeta=\hat{\scR}(\phi)$ and the $\scG$-flow $\phi=\hat{\scG}(\zeta)$ respectively define the encoding and decoding maps that connect the field $\phi$ in one-dimensional flat space to the field $\zeta$ in two-dimensional hyperbolic space, which explicitly realize the holographic duality in quantum gravity.

\subsection{Realization of Bijective Transformation} 

To optimize the renormalization transformation $\hat{\scR}$, one relies on the construction of a trainable bijective map to model $\hat{\scR}$. Machine learning community has provided several realizations of trainable bijective maps, including real NVP\cite{Dinh2016Density} and neural ODE\cite{Chen2018Neural}. In the following, we will focus on the neural ODE realization, as it can capture multi-modular features better than real NVP, which is more suitable for processing sequences of discrete objects.

\subsubsection{Neural ODE}
\label{Neural ODE}

Each single-step renormalization transform $(\phi',\zeta')=\hat{\scR}(\phi)$ can be realized by an ordinary differential equation (ODE). Starting from $\phi(0)=\phi$, first evolve $\phi(t)$ from $t=0$ to $t=1$ following
\begin{equation}\label{eq:ODE}
\frac{\dd\phi(t)}{\dd t}=f_\theta(\phi(t),t),
\end{equation}
where $f_\theta$ is a trainable function (realized as a neural network) parameterized by neural network parameters $\theta$. Then split the result as $\phi(1)=(\phi',\zeta')$ to obtain $\phi'$ and $\zeta'$. $t$ is considered as an auxiliary time. The inverse transformation is simply given by the time-reversal evolution, therefore the mapping is indeed bijective as desired. 

Apart from the transformation, the log Jacobian determinant of $\hat{\scR}$ can also be evaluated. Based on the ODE in \eqnref{eq:ODE}, one have
\begin{equation}
\log\Big\vert\det\frac{\partial\phi(t+\dd t)}{\partial\phi(t)}\Big\vert = \Tr\Big(\frac{\partial f_\theta(\phi(t),t)}{\partial \phi(t)}\Big)\dd t,
\end{equation}
which can be integrated to
\begin{equation}
\log\Big\vert\det\frac{\partial\hat{\scR}(\phi)}{\partial\phi}\Big\vert = \int_{0}^{1}\Tr\Big(\frac{\partial f_\theta(\phi(t),t)}{\partial \phi(t)}\Big)\dd t.
\end{equation}
Given that $\hat{\scG}:=\hat{\scR}^{-1}$, its log Jacobian determinant is simply given by a negation,
\begin{equation}
\log\Big\vert\det\frac{\partial\hat{\scG}(\phi',\zeta')}{\partial(\phi',\zeta')}\Big\vert = -\int_{0}^{1}\Tr\Big(\frac{\partial f_\theta(\phi(t),t)}{\partial \phi(t)}\Big)\dd t,
\end{equation}
which will be useful for the evaluation of the coupling action in \eqnref{eq:S_PhiZeta}.

\subsubsection{Locality and Translational Symmetry}

It is possible to design the ODE function $f_\theta$ to respect locality and translational symmetry. The idea is to realize $f_\theta$ using layers of convolutional neural networks (CNN) with finite kernel followed by element-wise activations. 

\subsection{Objective Function}

The objective is to train the generative model, such that the model distribution $p_{\Phi}(\phi)$ matches the data distribution $p_\text{dat}(\phi)$ as much as possible. The objective can be achieved by minimizing the Kullback-Leibler (KL) divergence
\begin{equation}
\begin{split}
\scL&=D_\text{KL}(p_\text{dat}\Vert p_{\Phi})=\int\scD\phi\;p_\text{dat}(\phi)\log\Big(\frac{p_\text{dat}(\phi)}{p_{\Phi}(\phi)}\Big)\\
&=\mathop{\dsE}_{\phi\sim p_\text{dat}}S_{\Phi}(\phi)-H(p_\text{dat}),
\end{split}
\end{equation}
where $S_{\Phi}(\phi)=-\log p_{\Phi}(\phi)$ is the model action (as the negative log-likelihood), and $H(p_\text{dat})=-\int\scD\phi\; p_\text{dat}(\phi)\log p_\text{dat}(\phi)$ is the data entropy. As the data entropy $H(p_\text{dat})$ is independent of the model parameter, it can be dropped from the loss function $\scL$. Therefore the loss function is essentially the ensemble average of the model action on the dataset. By minimizing the average action, the ODE function $f_\theta$ in each RG transformation will get trained. Upon convergence, the algorithm will find the optimal invertible RG flow that maps the (presumably) strongly coupled original field $\phi$ on the holographic boundary to the weakly coupled irrelevant field $\zeta$ in the holographic bulk. 

\subsection{Summary of the Algorithm} \label{Algorithm}
Given a set of sequences from the data, the learning algorithm goes as follows.
\begin{enumerate}
\item For each given sequence $a=(a_1,a_2,\cdots)$, represent each object $a_i$ in the sequence as a vector $\phi_i=E(a_i)\in\dsR^n$. Denote the sequence of vectors as a vector field $\phi=(\phi_1,\phi_2,\cdots)\in\mathsf{Map}(I,\dsR^n)$.

\item Starting with $\phi^{(0)}=\phi$, apply the renormalization transformation iteratively, 
\begin{equation*}
(\phi^{(k)},\zeta^{(k)})=\hat{\scR}^{(k)}(\phi^{(k-1)}),
\end{equation*}
for $K=\log_2|I|$ steps (until all relevant fields are eliminated).

\begin{enumerate}
\item Each step of the transformation is implemented by solving an ODE
\begin{equation*}
\frac{\dd\phi^{(k-1)}(t)}{\dd t}=f^{(k)}_\theta(\phi^{(k-1)}(t),t),
\end{equation*}
starting from the initial condition $\phi^{(k-1)}(0)=\phi^{(k-1)}$, integrating from $t=0$ to $t=1$, and then splitting the final result into $\phi^{(k-1)}(1)=(\phi^{(k)},\zeta^{(k)})$.
\item While solving the ODE, simultaneously integrate along the time evolution to obtain the coupling action
\begin{equation*}
S^{(k)}_{\Phi\Zeta}(\phi^{(k)},\zeta^{(k)}) = -\int_{0}^{1}\Tr\Big(\frac{\partial f^{(k)}_\theta(\phi^{(k-1)}(t),t)}{\partial \phi^{(k-1)}(t)}\Big)\dd t.
\end{equation*}
\end{enumerate}
\item Starting from the initial condition $S_{\Phi}^{(K)}=0$, collect the action in the reverse order (along the generation flow)
\begin{equation*}
S_{\Phi}^{(k-1)}(\phi^{(k-1)})=S_{\Phi}^{(k)}(\phi^{(k)})+ S_{\Zeta}^{(k)}(\zeta^{(k)})+S^{(k)}_{\Phi\Zeta}(\phi^{(k)},\zeta^{(k)}),
\end{equation*}
where $S_{\Zeta}^{(k)}(\zeta^{(k)})$ is given by
\begin{equation*}
S_{\Zeta}^{(k)}(\zeta^{(k)})=\frac{1}{2}\sum_{j\in J^{(k)}}\Vert\zeta^{(k)}_{j}\Vert^2.
\end{equation*}
The resulting total action will be denoted as $S_{\Phi}(\phi):=S_{\Phi}^{(0)}(\phi^{(0)})$.
\item Train the model to minimize the loss function
\begin{equation*}
\scL=\mathop{\dsE}_{\phi\sim p_\text{dat}}S_{\Phi}(\phi).
\end{equation*}
\end{enumerate}

\subsection{Potential Applications and Advantages} After training, the model could potentially be used for the following tasks.
\begin{itemize}
\item Inference of hierarchical latent representation. Using $\zeta=\hat{\scR}(\phi)$, one can infer the hierarchical latent representation $\zeta$ of any sequence encoding $\phi$. The high-level representations ($\zeta^{(k)}$ with a large $k$) can be viewed as the encoding of the entire sequence, which can be used in downstream tasks like classification and translation.

\item Likelihood estimation. Using $S_{\Phi}(\phi)$, one can estimate the probability density $p_{\Phi}(\phi)\propto \exp(-S_{\Phi}(\phi))$  for any field configuration $\phi$. This will be useful for anomaly detection.

\item Sample generation. As a generative model, new samples can be generated by first sampling $\zeta$ in the hyperbolic space and then transforming to $\phi=\hat{\scG}(\zeta)$ using the generation flow, which may find applications in completing missing objects in a sequence.
\end{itemize}

The propose algorithm is advantageous in the following aspects.
\begin{itemize}
\item Disentangled features in scales. The optimal RG flow distills features at different scales, allowing the model to capture the long-range and multi-scale correlation in the sequential data. The features are automatically arranged in a hyperbolic spaces, making it easy to access/control.

\item Efficient inference/generation. The hierarchical and iterative approach enables the model to infer latent fields or generate original fields in $\Theta(N)$ complexity (given the sequence length $N$), which is superior compared to the $\Theta(N^2)$ complexity of transformer-based approaches, especially when the sequence is long.

\item Ability to process hierarchical structure. The renormalization transformation can progressively extract coarse-grained features from fine-grained features, making it capable of capturing global features (such as the parity of bit strings). In comparison, as shown in Ref.\,\cite{Hahn2019Theoretical}, self-attention-based models can not efficiently model hierarchical structures, unless the number of layers/heads increases with sequence length.

\end{itemize}

\subsection{Recovering Conventional RG by Integrating out Irrelevant Fields.}\label{sec:conventional RG} Finally, we would like to comment that the invertible renormalization can fall back to the conventional renormalization by integrating out irrelevant fields. Recall \eqnref{eq:Rk} that in each step of the invertible renormalization transformation, the original (fine-grained) field $\phi^{(k-1)}$ is separated into the relevant $\phi^{(k)}$ and irrelevant $\zeta^{(k)}$ fields by $(\phi^{(k)},\zeta^{(k)})=\hat{\scR}^{(k)}(\phi^{(k-1)})$. The invertible renormalization $\hat{\scR}^{(k)}$ can be downgraded to a non-invertible renormalization $\scR^{(k)}$ by a forgetful map which forgets about the irrelevant field $\zeta^{(k)}$, such that $\phi^{(k)}=\scR^{(k)}(\phi^{(k-1)})$ only transforms the the fine-grained field $\phi^{(k-1)}$ to the coarse-grained field $\phi^{(k)}$.

According to \eqnref{eq:recursive action}, the actions are related by
\begin{equation}
S_{\Phi}^{(k-1)}(\phi^{(k-1)})=S_{\Phi}^{(k)}(\phi^{(k)})+ S_{\Zeta}^{(k)}(\zeta^{(k)})+S^{(k)}_{\Phi\Zeta}(\phi^{(k)},\zeta^{(k)}),
\end{equation}
where the irrelevant field $\zeta^{(k)}$ is massive, and is described by the Gaussian action 
$S_{\Zeta}^{(k)}(\zeta^{(k)})=\frac{1}{2}\Vert\zeta^{(k)}\Vert^2$ as in \eqnref{eq:SZ}. Because $\zeta^{(k)}$ represents the high-energy modes that should be integrated out under renormalization, one can argue that the fluctuation of $\zeta^{(k)}$ can be treated perturbatively due to its large mass, which justifies the expansion of the action around $\zeta^{(k)}\to 0$,
\begin{equation}
\begin{split}
S_{\Phi}^{(k-1)}(\phi^{(k-1)})&\simeq S_{\Phi}^{(k)}(\phi^{(k)})+S^{(k)}_{\Phi\Zeta}(\phi^{(k)},0)+\zeta^{(k)}\cdot\partial_{\zeta^{(k)}}S^{(k)}_{\Phi\Zeta}(\phi^{(k)},0)\\
&+\frac{1}{2}\Vert\zeta^{(k)}\Vert^2+\frac{1}{2}\zeta^{(k)}\cdot\partial_{\zeta^{(k)}}\partial_{\zeta^{(k)}}S^{(k)}_{\Phi\Zeta}(\phi^{(k)},0)\cdot \zeta^{(k)}+\cdots.
\end{split}
\end{equation}
As the approximate action is quadratic in $\zeta^{(k)}$, one can perform a Gaussian integration for $\zeta^{(k)}$, under which the action becomes
\begin{equation}\label{eq:effective action}
S_{\Phi}^{(k-1)}(\phi^{(k-1)})= S_{\Phi}^{(k)}(\phi^{(k)})+S^{(k)}_{\Phi\Zeta}(\phi^{(k)},0)+\frac{1}{2}\big(\partial_{\zeta^{(k)}}^2 S^{(k)}_{\Phi\Zeta}(\phi^{(k)},0)-\big(\partial_{\zeta^{(k)}}S^{(k)}_{\Phi\Zeta}(\phi^{(k)},0)\big)^2\big).
\end{equation}
Therefore one can define the renormalization transformation $\bar{\scR}^{(k)}$ on the action via $S_{\Phi}^{(k)}=\bar{\scR}^{(k)}(S_{\Phi}^{(k-1)})$, in correspondence to the field renormalization $\phi^{(k)}=\scR^{(k)}(\phi^{(k-1)})$. Based on \eqnref{eq:effective action}, the explicit form of the renormalization operator $\bar{\scR}$ can be given
\begin{equation}
\bar{\scR}(S_{\Phi})=S_{\Phi}-S_{\Phi\Zeta}-\frac{1}{2}(\partial_{\zeta}^2 S_{\Phi\Zeta}-(\partial_{\zeta}S_{\Phi\Zeta})^2),
\end{equation}
such that
\begin{equation}
\bar{\scR}(S_{\Phi})(\phi) = S_{\Phi}(\scR(\phi)),
\end{equation}
which reproduces the pullback construction of the action renormalization. If one further define the infinitesimal generator of $\bar{\scR}$ as $\bar{\mathfrak{r}}=\log\bar{\scR}$, the renormalization flow can be expressed as a differential equation\cite{Polchinski1984Renormalization,Ma2020Constraints}
 $\partial_{\ell}S_{\Phi}=\bar{\mathfrak{r}}S_{\Phi}=-S_{\Phi\Zeta}-\frac{1}{2}(\partial_{\zeta}^2 S_{\Phi\Zeta}-(\partial_{\zeta}S_{\Phi\Zeta})^2).$
 
\section{Experiments on genomic sequences}

\subsection{Problem Overview}
Extracting the hidden information of genomic sequences has been a critical subject in biological research, with relevance to epidemiology, immuniology, protein design and many other subfields. With its great similarity to the natural language processing problems, there are numerous studies on applying machine learning techniques to extract information from genomic sequences. Various machine learning architectures include word2vec \cite{du2019gene2vec}\cite{wu2021replication}, bidirectional long short-term memory\cite{hie2021escape}, transformer\cite{ji2021transformer} etc. However while the existing algorithms can provide a single-gene level embedding, they do not provide a canonical sequence embedding and the hierarchical information is not clear from the natural language models. Thus we apply the renormalization group idea from the previous sections to the genomic sequence representation problem, where the hierachial structure provides the biological information at different energy levels, i.e. the deeper layer can capture the longer correlation in the sequence, and thus provide a canonical embedding of the sequence with the deepest layer. We take the Influenza HA amino acid sequences as an example\footnote{can be downlaoded from the “Protein Sequence Search'” section of \url{https://www.fludb.org}}, where the sequences are regarded as one-dimensional lattices as described in Sec.~\ref{1dlattice}. Below shows samples of the sequence data, there is clear global features embedded as one can see the similarities between different sequences.

\begin{figure}[htbp]
\begin{center}
\includegraphics[width=0.85\columnwidth]{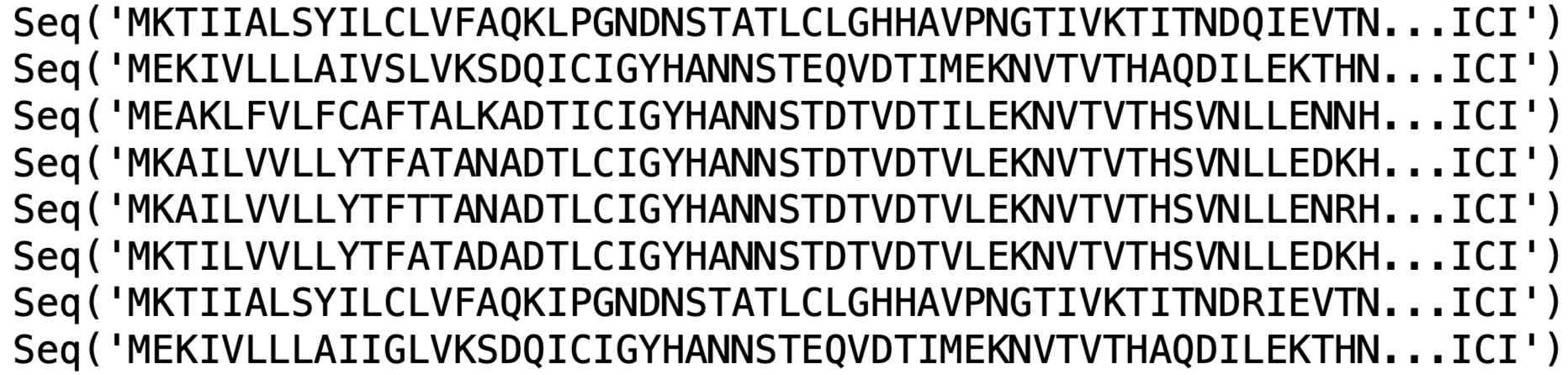}
\caption{Sample Influenza HA amino acid sequences.}
\label{fig:Influenza_seq}
\end{center}
\end{figure}

\subsection{Single Amino Acid Distribution Learning}
Before proceeding with the sequence into the RG-scheme, we need to verify that local features are efficiently learned. Thus we first look at the single amino acid distribution learning. As shown in Fig.~\ref{fig:Influenza_seq}, at a fixed location $i$ among the sequences, there is a discrete distribution labeled by amino acid, we pick up that amino acid from each sequence. Then each sample is labeled by $a=(a_i)$, where $a_i\in A$ represents the single amino acid. We apply the pre-trained single-amino-acid level embedding $E: A\rightarrow \mathbb{R}^n$ from Ref.~\cite{hie2021escape} where $n=20$. After embedding, the boundary field is $\phi = (\phi_i)$, where $\phi_i=E(a_i)\in \mathbb{R}^n$. To remove the difficulty in transforming the boundary discrete distribution to the bulk uncorrelated continuous Gaussian distribution, we add a small randomness on the boundary field, i.e. $\overline{\phi_i} = \phi_i + \epsilon$, where $p_Z(\epsilon)$ takes the standard normal distribution with small variance. For simplicity, in the following we still use $\phi_i$ to denote the fields with small randomness.

With this setup, we train a neural ODE model to realize the bijective transformation between the data distribution to a Gaussian distribution as described in Sec.~\ref{Neural ODE}. To further speed up the training process, we have added the Jacobian and Kinetic regularization to find an optimal bijective map as in Ref.~\cite{finlay2020train}. The input data is the 4th single amino acid from 1000 Influenza HA sequences.

\begin{figure}[htbp]
\begin{center}
\includegraphics[width=0.25\columnwidth]{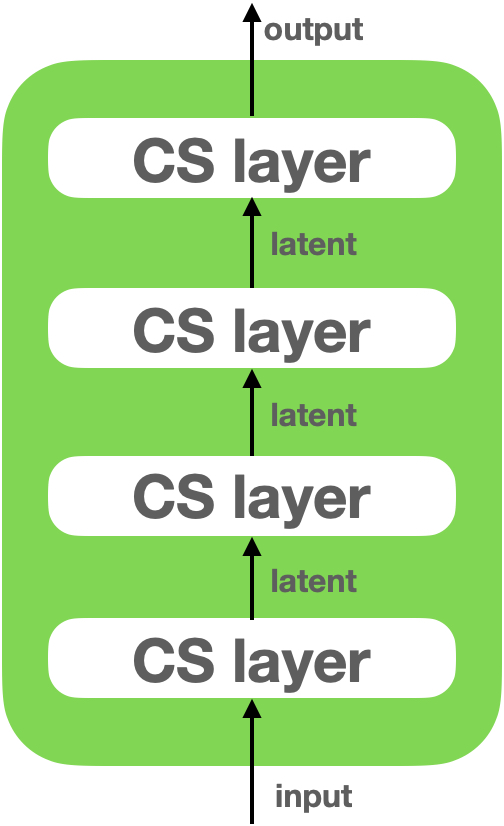}
\caption{Neural ODE structure. The input is the single amino acid vector representation $\phi\in \mathcal{R}^n$ with $n=20$. The hidden layers are concatsquash(CS) layers. The latent varible has dimension $64$ and the output has dimension $n=20$.}
\label{fig:NODE_structure}
\end{center}
\end{figure}

The ODE transformation $f_\theta(x,t)$ is constructed by a feed forward network with 4 sequential hidden layers as shown in Fig.~\ref{fig:NODE_structure}. The hidden layers are the concatsquash(CS) layers defined in Ref.~\cite{grathwohl2018ffjord}:
\begin{equation}
f_{CS}(x,t) = W_0 x \cdot \sigma(W_1 t) + W_2 t
\end{equation}
where $W_0$, $W_1$, $W_2$ are parameter matrices with shape $(d_{o},d_{i})$, $(d_{o},1)$, $(d_{o},1)$ respectively, $d_i$ and $d_o$ are input and output dimensions. We can consider $x$ as a $d_i$-dimenional vector, and concate the time variable, then $f_{CS}: \mathbb{R}^{d_i+1}\rightarrow \mathbb{R}^{d_o}$. The hyperbolic tangent activation functions are applied after the first three CS layers. Then the model is trained such that $f_{\theta}(x,t)$ describes the flow from the data to a Gaussian variable.

\begin{figure}[htbp]
     \centering
     \begin{subfigure}[b]{0.45\textwidth}
         \centering
         \includegraphics[width=\textwidth]{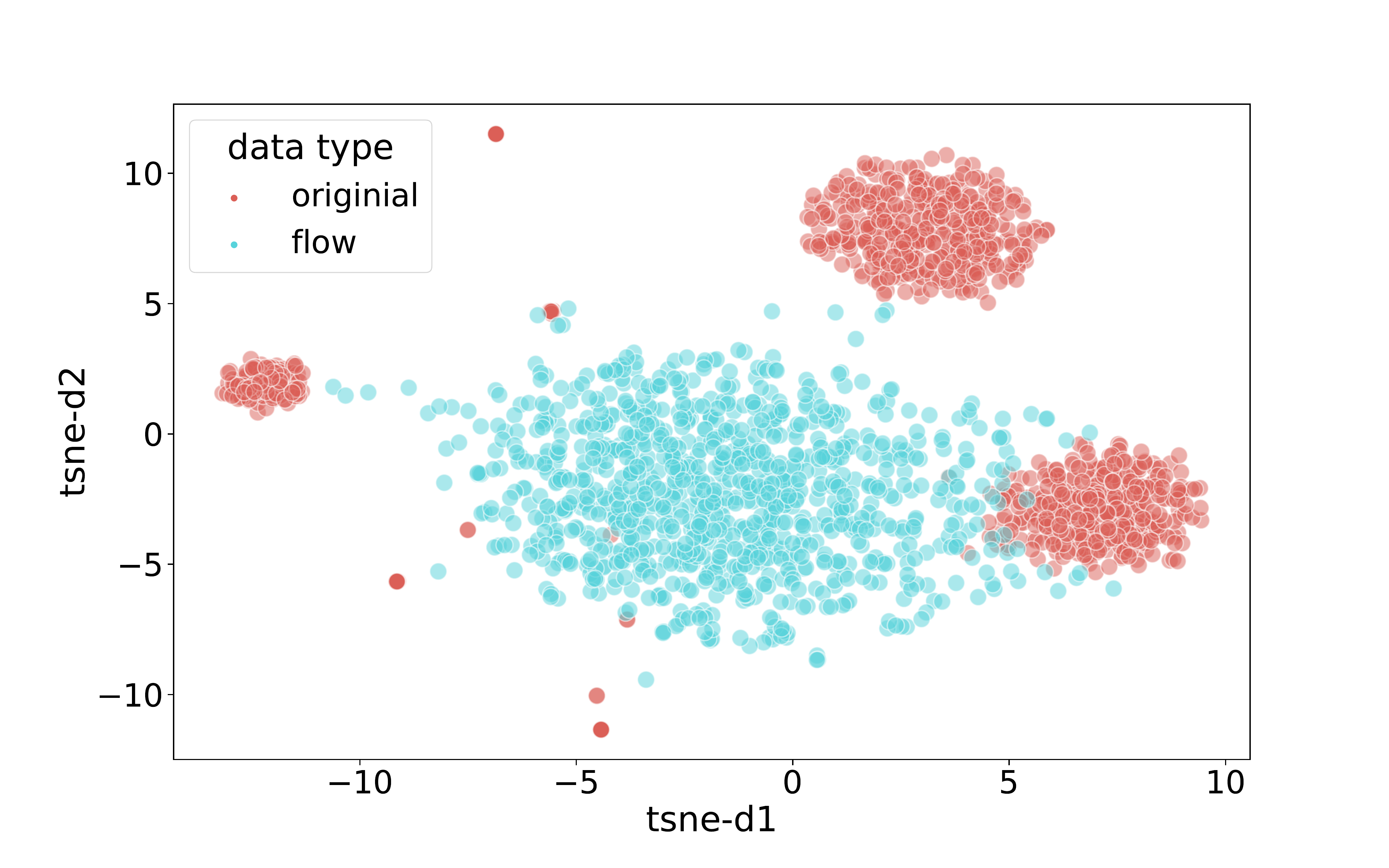}
         \caption{Initial distribution}
     \end{subfigure}
     \begin{subfigure}[b]{0.45\textwidth}
         \centering
         \includegraphics[width=\textwidth]{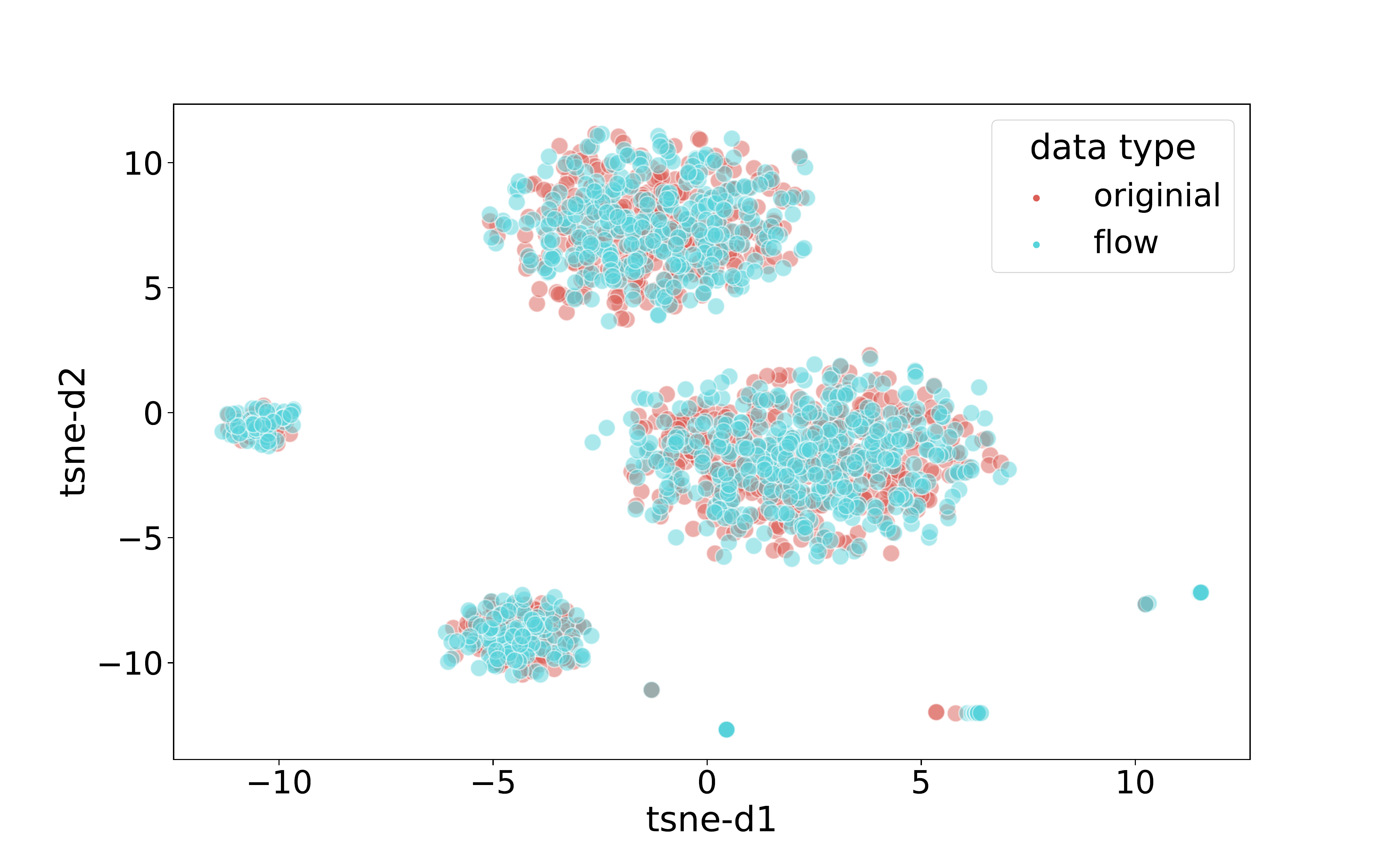}
         \caption{Distribution after training}
     \end{subfigure}
        \caption{Training result on a single amino acid with t-SNE representation. (A): The original data distribution(red) has a multi-modular struture, while the initial flow model's distribution(blue) is standard normal. (B): After training with neural ODE, two distributions coincide.}
        \label{fig:single_amino_acid}
\end{figure}

As shown in Fig.~\ref{fig:single_amino_acid}, a 2-dimensional feature space can be obtained by applying the t-distributed stochastic neighbor embedding (t-SNE) algorithm\cite{JMLR:v9:vandermaaten08a} to the original data and the flow generated data embedding vectors. The flow generated data are obtained by taking the inverse transformation $\hat{\mathcal{R}}^{-1}$ from vectors with the Gaussian distribution. The original data distribution with the multi-modular feature can be perfectly captured after training. 

\subsection{Amino Acid Sequence Distribution Learning}
\begin{figure}[htbp]
\begin{center}
\includegraphics[width=0.5\columnwidth]{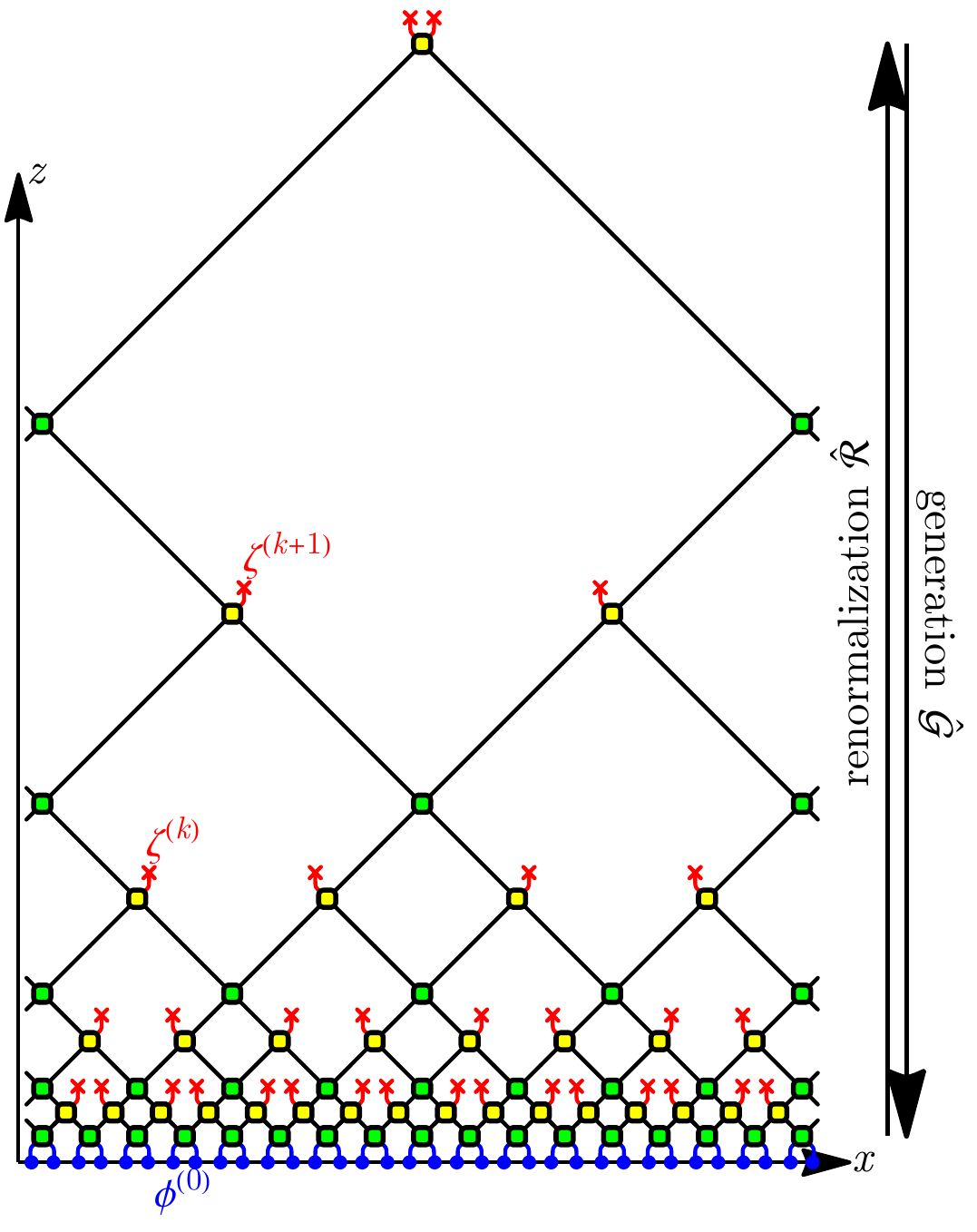}
\caption{An illustration of the mera structure with kernel $l=2$. Green blocks are disentangler blocks, yellow blocks are decimator blocks. After each decimator layer, half of the fields are redefined as bulk fields $\zeta$. }
\label{fig:MERA}
\end{center}
\end{figure}

To train on sequences, we adapt the hierachical RG scheme described in Sec.~\ref{Hierarchical Structure}. As discussed in the previous section, the input sequence is represented as labels $a = (a_1,a_2,\cdots, a_I)$ with the cardinality $I$ denotes the length of the sequence. With the pretrained embedding $\phi_i=E(a_i)\in \mathbb{R}^n$, the boundary fields are represented as $\phi = (\phi_1,\phi_2,\cdots, \phi_I)$. Thus we have the initial boundary fields $\phi^{(0)}=\phi$, we can run the renormalization flow using Eq.~\ref{eq:R flow}. On the other hand, the generation flow Eq.~\ref{eq:G flow} reconstruct the original field. Following the notation of MERA networks, each renormalization transformation layer consists of a disentangler layer and a decimator layer:
\begin{equation}
\hat{\scR}^{(k)} = \hat{\scR}^{\text{dec},(k)}\circ \hat{\scR}^{\text{dis},(k)}
\end{equation}
where a disentangler layer disentangles the local correlations and a decimator layer separate the decimated fields out as the bulk fields. In Fig.~\ref{fig:MERA}, we show an illustration of the model structure with green blocks as disentanglers and yellow blocks as decimators. Each block is a bijective transformation with the neural ODE structure as in Fig~\ref{fig:NODE_structure}. We can further explicitly write down the transformation equations: the covering length of a disentangler or a decimator is defined as the kernel length $l$. Then there are $\frac{I}{2^k l}$ blocks in the $k$-th layer. For the $m$-th block in the $k$-th layer, where $m\in \lbrace 0,\cdots,\frac{I}{2^k l} -1\rbrace$, the transformation is given by
\begin{equation}
\begin{split}
&(\psi^{(k)}_{2^k(ml+a)})_{a\in\lbrace 1,\cdots,l\rbrace} = \hat{\mathcal{R}}^{\text{dis},(k)}_m((\phi^{(k)}_{2^k(ml+a)})_{a\in \lbrace 1,\cdots,l\rbrace})\\
&(\phi^{(k+1)}_{2^k(ml+l/2+a)})_{a\in\lbrace 1,\cdots,l\rbrace} = \hat{\mathcal{R}}^{\text{dec},(k)}_m((\psi^{(k)}_{2^k(ml+l/2+a)})_{a\in \lbrace 1,\cdots,l\rbrace})
\end{split}
\end{equation}
Half of the resulting fields are redefined as the bulk fields: $\zeta^{(k+1)}_{2^{k+1}(ml+a)}\coloneqq\phi^{(k+1)}_{2^{k+1}(ml+a)}$. Here we have chosen a scheme that after each layer, every other existing fields are redefined as new bulk fields. Since there are position-dependent features among the sequences, to respect the local features of the sequence, we take independent block transformations as they are labeled by both layer index $k$ and block index $m$. With this setup, we train on the objective $\scL=\mathop{\dsE}_{\phi\sim p_\text{dat}}S_{\Phi}(\phi)$ as described in Sec.~\ref{Algorithm}. 

\begin{figure}[htbp]
     \centering
     \begin{subfigure}[b]{0.45\textwidth}
         \centering
         \includegraphics[width=\textwidth]{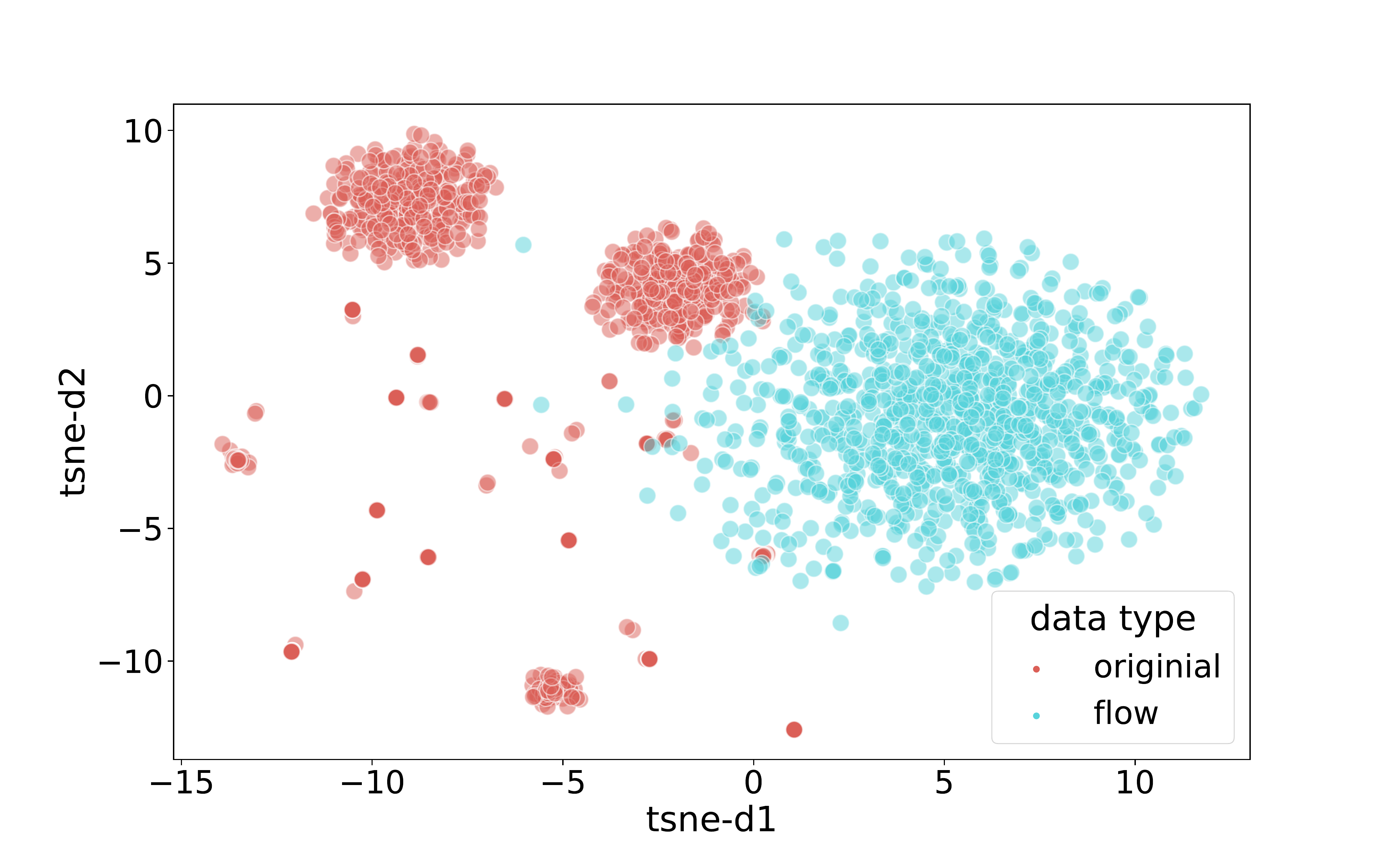}
         \caption{Initial distribution}
     \end{subfigure}
     \begin{subfigure}[b]{0.45\textwidth}
         \centering
         \includegraphics[width=\textwidth]{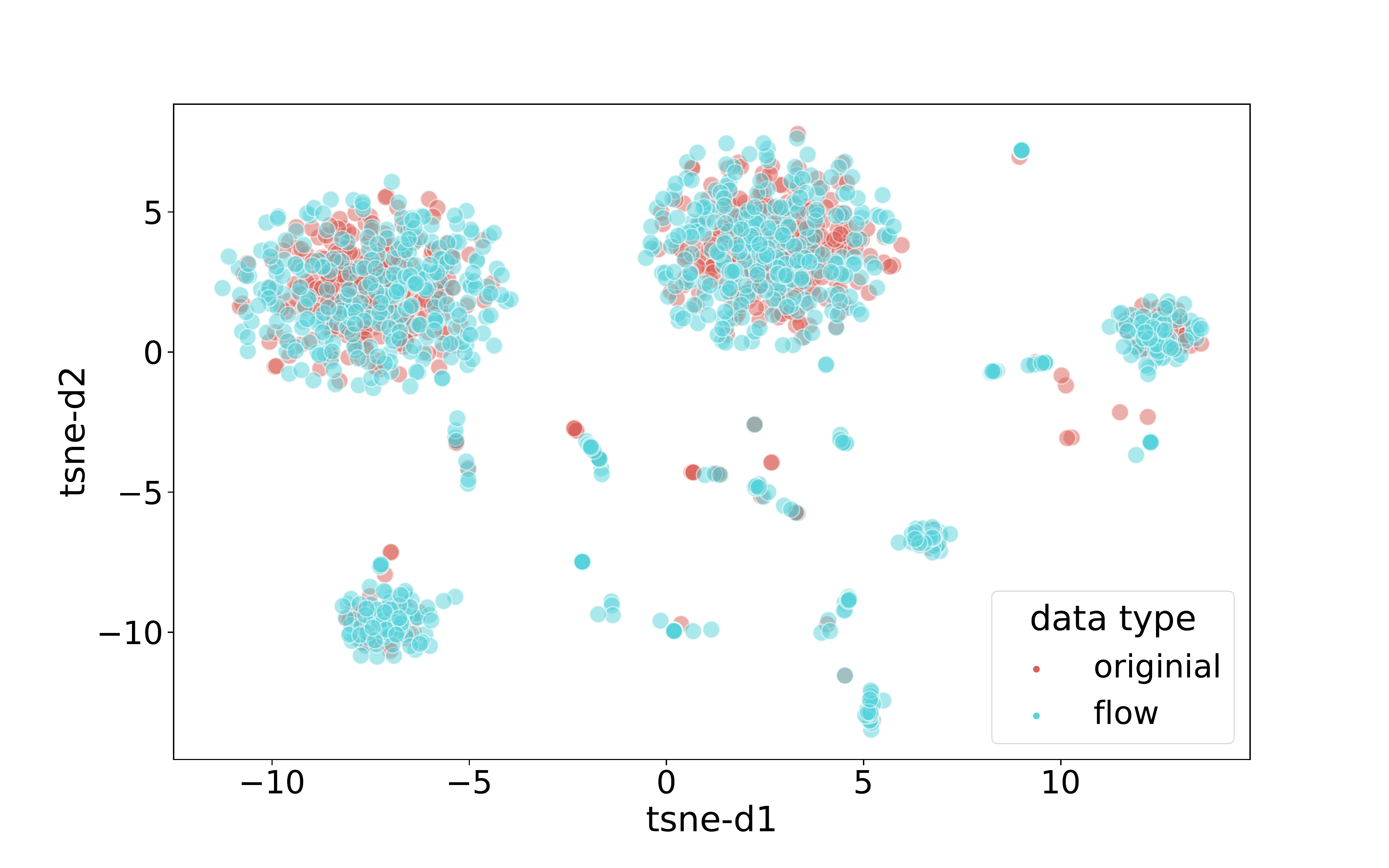}
         \caption{Distribution after training}
     \end{subfigure}
        \caption{Training result on a length 4 amino acid sequence dataset with t-SNE representation. (A): The original data distribution(red) has a multi-modular struture, while the initial flow model's distribution is standard normal(blue). (B): After training RG scheme with neural ODE blocks, two distributions concide.}
        \label{fig:4_amino_acid}
\end{figure}

\begin{figure}[htbp]
     \centering
     \begin{subfigure}[b]{0.45\textwidth}
         \centering
         \includegraphics[width=\textwidth]{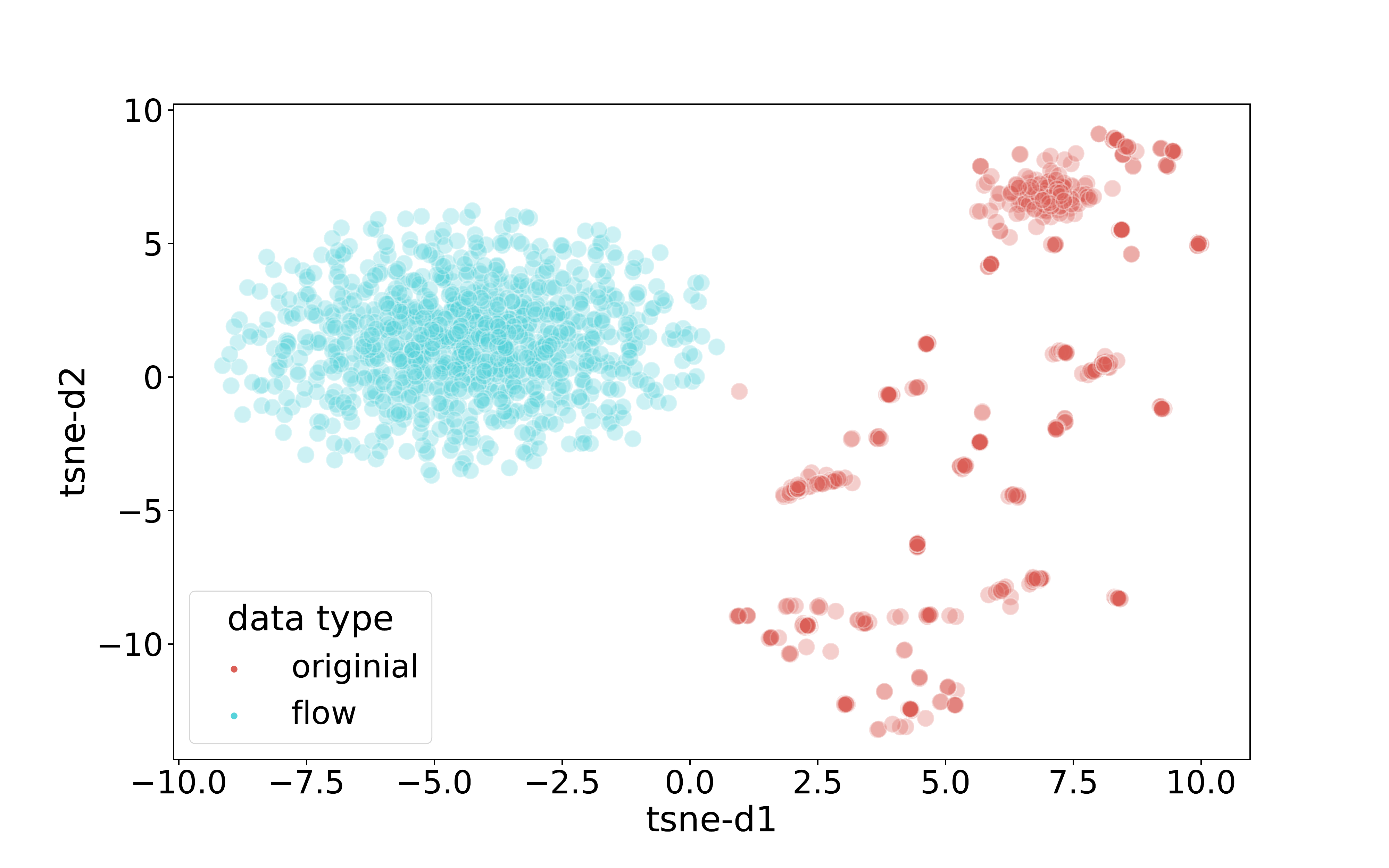}
         \caption{Initial distribution}
     \end{subfigure}
     \begin{subfigure}[b]{0.45\textwidth}
         \centering
         \includegraphics[width=\textwidth]{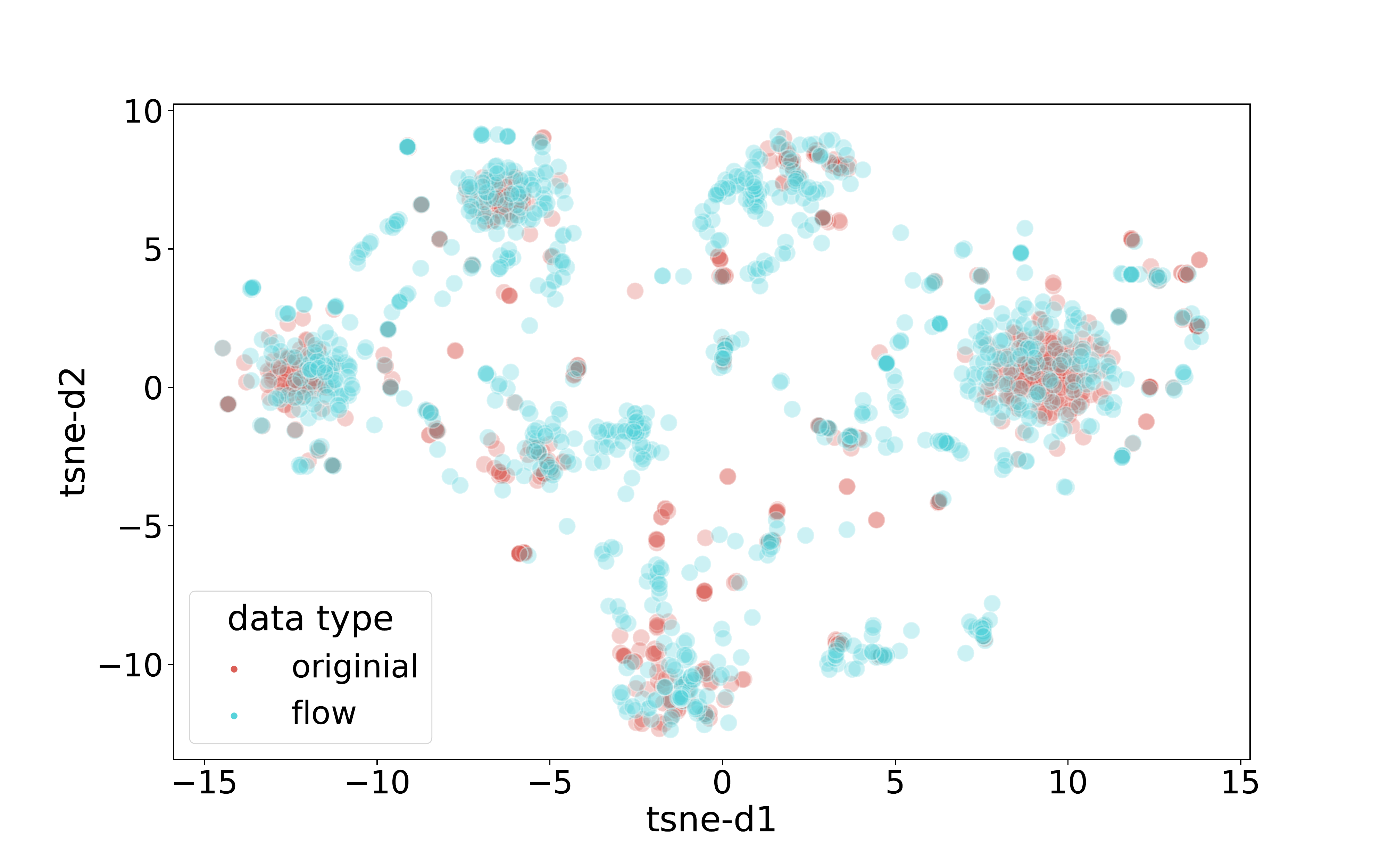}
         \caption{Distribution after training}
     \end{subfigure}
        \caption{Training result on a length 16 amino acid sequence dataset with t-SNE representation. (A): The original data distribution(red) has a multi-modular struture, while the initial flow model's distribution is standard normal(blue). (B): After training RG scheme with neural ODE blocks, two distributions concide.}
        \label{fig:16_amino_acid_4500}
\end{figure}

In Fig.~\ref{fig:4_amino_acid} and Fig.~\ref{fig:16_amino_acid_4500}, we show the result when $I=4$, $l=2$ and when $I=16$, $l=4$ with the same set of data in the previous section. To compare the joint distribution, we concatenate the vector embeddings of the 4 and 16 amino acids for each sequence and train the t-SNE algorhim with these concated vectors. We also computed the normalized logarithmic probability defined as $\log_n p = \log p /(nI)$ with $n$ the embedding dimension and $I$ the sequence length. The numbers in the parentheses are normalized logarithmic probability before training. Both results shows that the original data joint distribution can be captured using the RG-scheme with local neural ODE blocks.

\begin{table}[htbp]
\centering
\begin{tabular}{ccc}
	\toprule
	& \multicolumn{2}{c}{normalized log prob}\\
	\cmidrule{2-3}
	&  seq length $ =4$ &  seq length $ =16$\\
	\midrule
	original data & 1.52(0.45) & 1.52(0.25)\\
	generated data & 1.50(0.19) & 1.50(0.19)\\
	\bottomrule
\end{tabular}
\caption{\label{tab:log_prob}Average normalized logarithmic probability from original data and generated data with sequence length 4 and 16. }
\end{table}

With training on the full sequence, one can have hierarchical information from each layer. This may give a natural way for escaping virus search. The shallow layers mainly capture the local information, while the deeper layers hold the global information. Escaping virus should have a good local fitness while mean a different content compared with the existing dataset. Then one can use this separation of information levels to design rules for escaping virus or train on a downstream classification task.

\subsection{Learning Viral Escape Mutation}
\begin{figure}[h]
\begin{center}
\includegraphics[width=0.65\columnwidth]{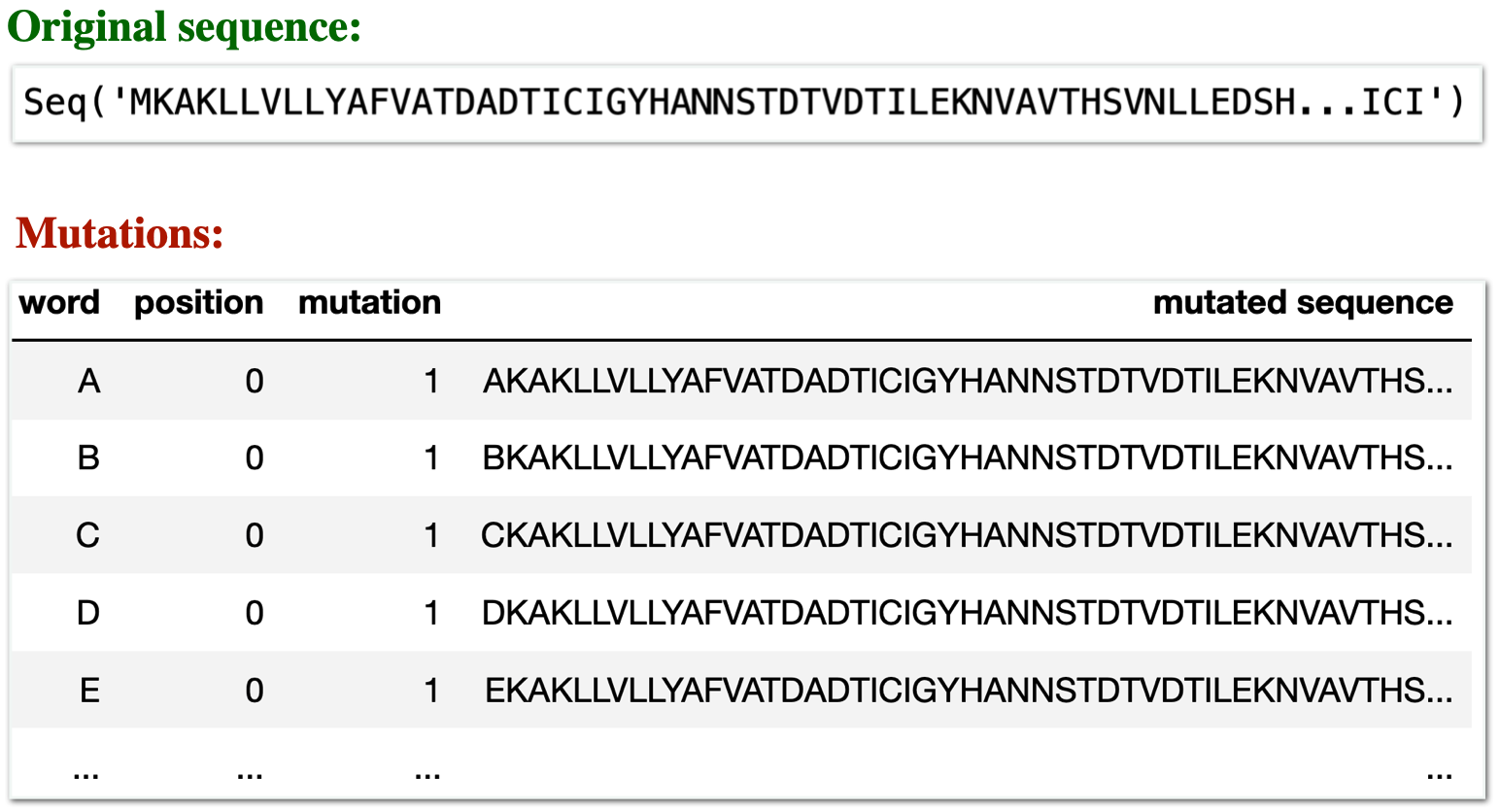}
\caption{Sample of possible mutations in the test sequence. First column "word" represents the mutant amino acid, "position" indicates the mutation position in the sequence.}
\label{fig:viral_esc}
\end{center}
\end{figure}

We conclude our investigation on learning protein sequences distribution by studying the predictive performance of viral escapes with our model. Viral escapes are those mutations in viral protein sequences that make them unrecognizable for human immune system. In other words, although they are still effective on human body and cause infection, the immune system does not flag the mutated sequence as a threat to body. Such mutations can be single or multiple, that is, only one or few amino acids can be instantly mutated, hence, identifying underlying patterns in viral escape mutations will be essential for viral vaccine development. As described in Ref.~\cite{hie2021escape}, in terms of language models, a viral sequence can be regarded as a textual data and a viral escape mutation is seen as a word change in a sentence that does changes the semantic of the sentence however the sentence is still grammatically meaningful. With this analogy, a viral escape is the one capable of making the immune system falsely flag the mutant as a harmless sequence (change in sentiment), while the mutant preserves the virus evolutionary structure (grammatically correct). Therefore, among all possible mutations in a viral sequence, we search for viral escapes which result in a high semantic change and high gramaticality in our model. Figure~\ref{fig:viral_esc} depicts an example of all possible mutations in the test sequence.

Following this idea (constrained semantic change search (CSCS)~\cite{hie2021escape}) we first train our model on a corpus of viral sequences in an unsupervised fashion, then take a given viral protein sequence with its known viral escape mutations and rank the mutations based on their gramaticality and semantic change. In our construction, the semantic change, caused by a mutation, is regarded as the change in the internal representation of the deepest layer in our construction before and after mutation happens. In other words, given the test sequence $a =(a_1, a_2, ,a_i,.., a_I)$ and its mutant counterpart $\bar{a} =(a_1, a_2, ,\bar{a}_i,.., a_I)$, a semantic change will be noted as $\Delta \zeta= |\zeta^{(K)}(a) -\zeta^{(K)}(\bar{a})|$ where $K$ indicates the deepest layer in the hierarchical structure of our model.

According to the CSCS objective, gramaticality can be defined as how probable is a mutation in $a$, i.e, the probability value that model evaluates for a mutation. With this definition, one natural definition of gramaticality is the conditional probability $p(\bar{a_{i}}|a)$ on the mutation $\bar{a_{i}}$ in the test protein~\cite{hie2021escape}. In our model however, the joint probability $p_{\Phi}(a)$ is the optimization objective which is used to evaluate the input gramaticality. Therefore the final score for each mutation is defined as:
\begin{equation}
Score \coloneqq \Delta \zeta + p(\bar{a}_i|a) 
\label{eq::score}
\end{equation}

\begin{figure}[htbp]
     \centering
     \begin{subfigure}[b]{0.45\textwidth}
         \centering
         \includegraphics[width=\textwidth]{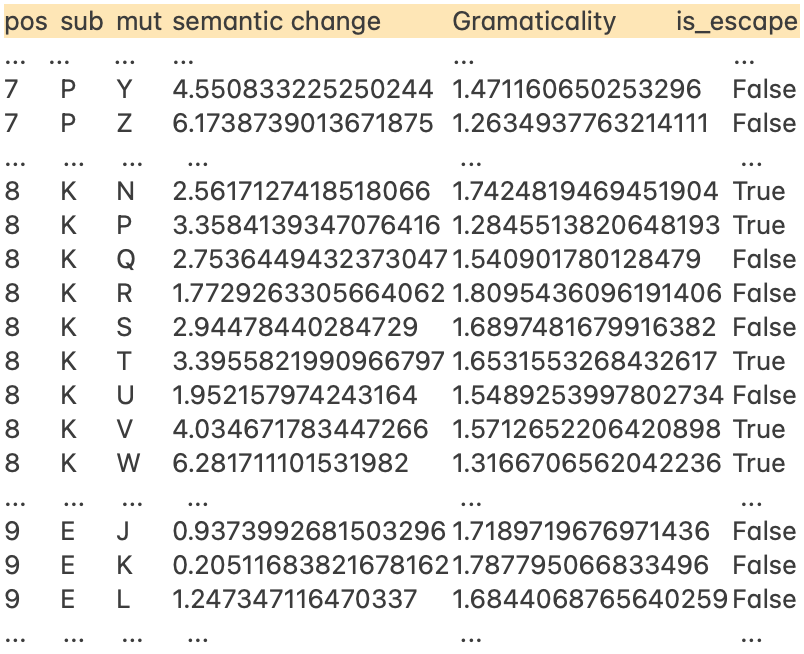}
         \caption{}
        \label{fig::deltap_z_table}
     \end{subfigure}
     \begin{subfigure}[b]{0.45\textwidth}
         \centering
         \includegraphics[width=\textwidth]{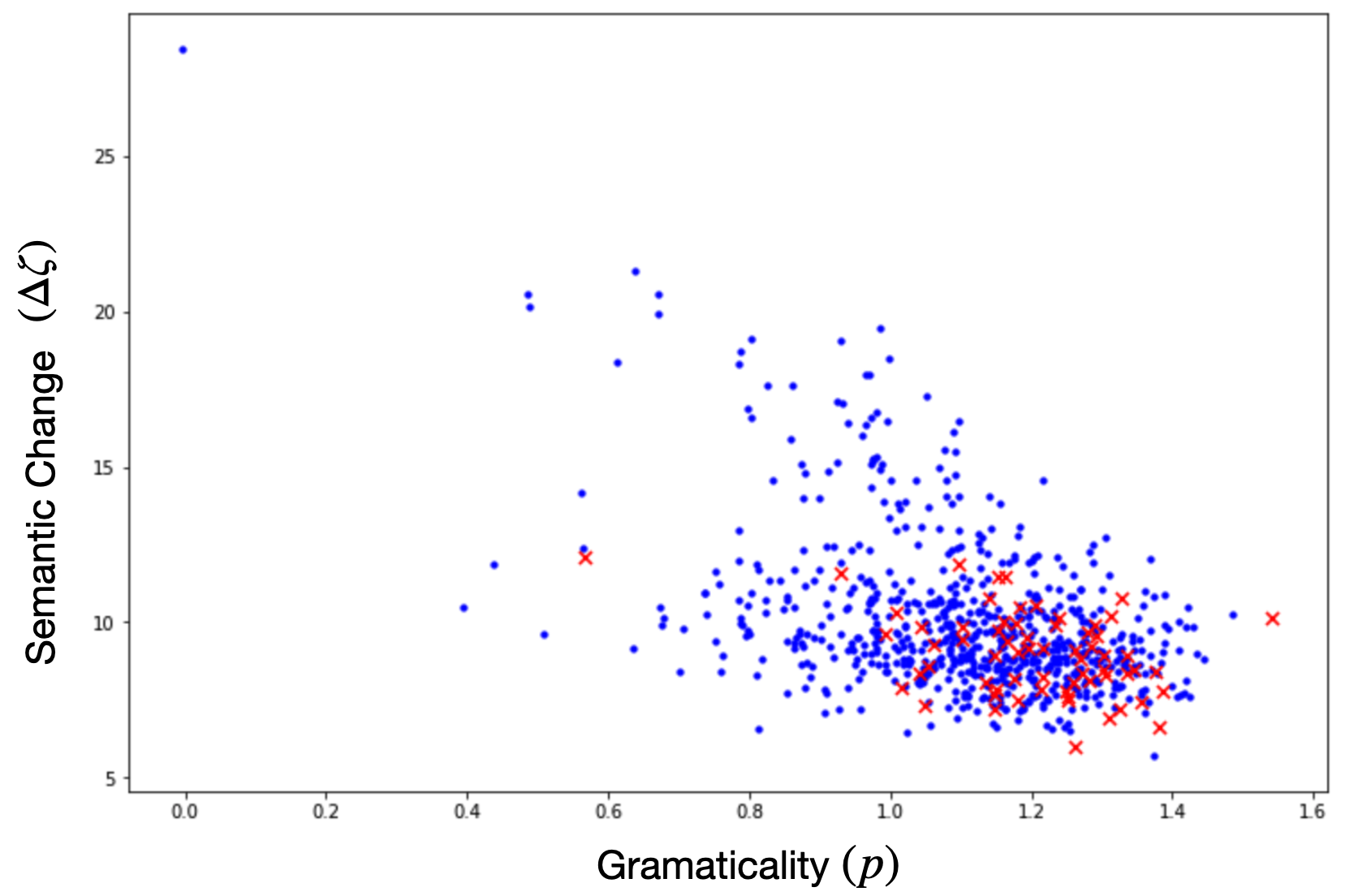}
         \caption{}
        \label{fig::deltap_z}
     \end{subfigure}
    \caption{(A) Table of possible mutations with their Gramaticality and Semantic change. In the list of columns, "pos" means the position in sequence to be mutated, "sub" refers the substitution word, "mut" is the mutant amino acid, and "is-escape" shows which substitution is a viral escape.(B) Gramaticality vs Semantic change of all mutations. Red points indicates the viral escape mutations. Note that the graph is not scaled.}     
\end{figure}

Note that throughout evaluating our model on viral escape mutations, we only consider single mutation in test data. We also keep the size of our samples to 32 amino acids, and with only 25 different amino acids as the building blocks of sequences, there will be 768 (24x32) possible mutations. Among those, a small subset will be viral escape mutations that is already given to us. For this experiment we used escape mutations dataset in \cite{doud2018single} that indicates 65 out of those 768 mutations are viral escapes. After calculating both sentiment change and grammatically of mutations, we ranked each mutation based on their ranking score in Eq.\ref{eq::score}. Mutations with highest ranking value will be considered as predicted viral escapes and consequently, lower rankings indicate less probable for a mutation to be a viral escape. Fig~\ref{fig::deltap_z_table} and Fig~\ref{fig::deltap_z} illustrate gramaticality and semantic change of all mutations including the viral escapes (red points) which clearly shows that viral escapes tend to show a high gramaticality.

\begin{figure}[htbp]
     \centering
     \includegraphics[width=0.5\textwidth]{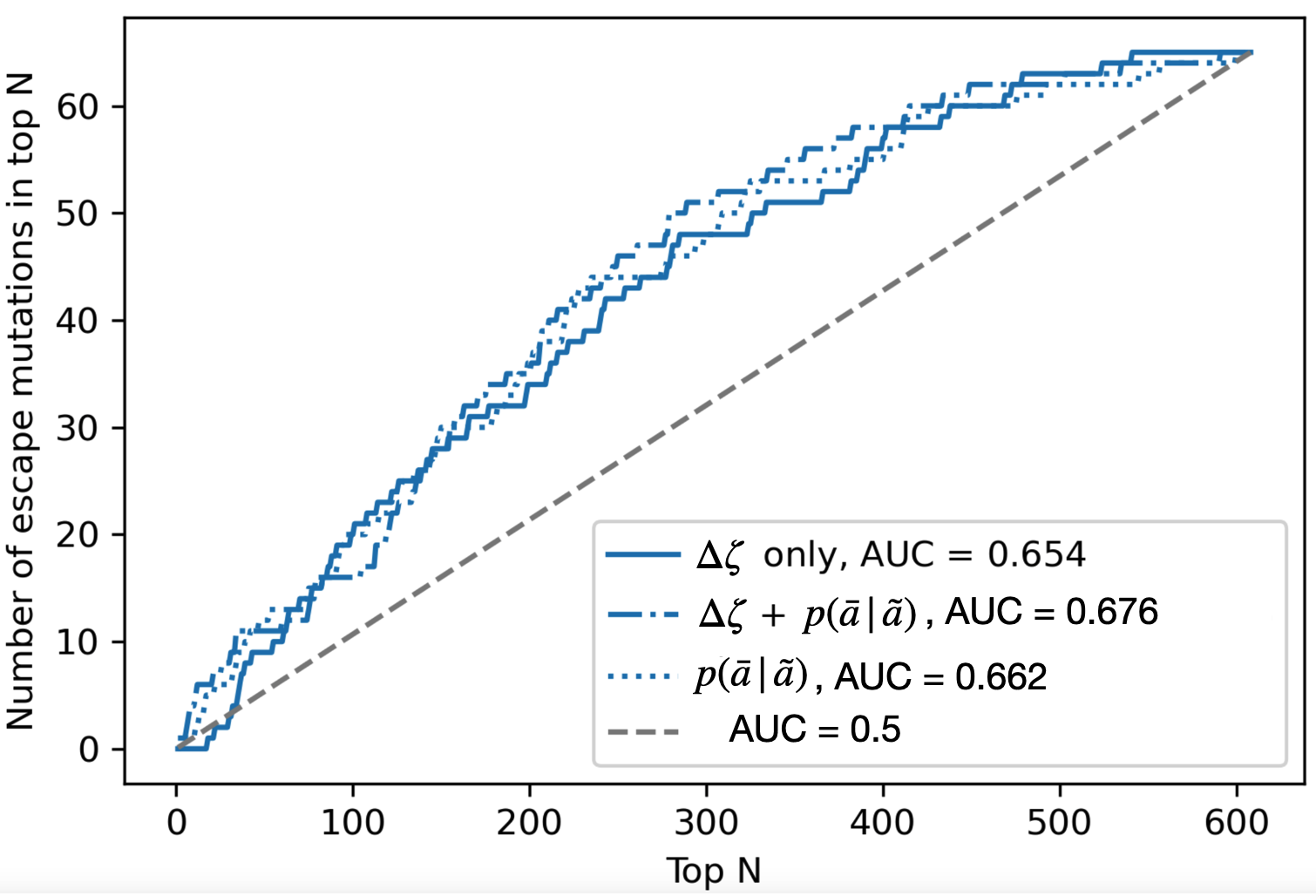}
    \caption{AUC graph of mutations in the test viral protein sequence with 32 amino acids. Gramaticality ($p$) and semantic change ($\Delta \zeta $) have equivalent contribution to the final AUC value.}     
    \label{fig::auc}    
\end{figure}

We also calculated the area under curve (AUC) of the ranking scores of mutations in Fig.~\ref{fig::auc}. Our results clearly indicate that both grammaticality and semantic change quantities have similar impact on the AUC value.

\newpage
\bibliographystyle{plain}
\bibliography{ref}

\noindent{\small{\tt{asheshmani@fas.harvard.edu, yzyou@physics.ucsd.edu, \\\\sd052fenber@gmail.com}, razizi@qgnai.com}

\end{document}